\documentclass[twoside]{article}

\usepackage[left=2cm, right=2cm, bottom=2cm, top=2cm]{geometry}
\usepackage[round]{natbib}
\usepackage[english]{babel}
\usepackage[utf8]{inputenc} % allow utf-8 input
\usepackage[colorlinks = true,
          				  urlcolor  = blue,
          				  citecolor = black,
          				  linkcolor=black,
           				  anchorcolor = black]{hyperref}
\usepackage{url}            % simple URL typesetting
\usepackage{booktabs}       % professional-quality tables
\usepackage{amsfonts, mathrsfs, amssymb}       % blackboard math symbols
\usepackage{nicefrac}       % compact symbols for 1/2, etc.
\usepackage{microtype}      % microtypography
\usepackage[export]{adjustbox}

\usepackage{graphicx}
\usepackage{subfigure,wrapfig}
\usepackage{dsfont}
\usepackage{amsmath}
\usepackage{amsthm}
\usepackage{stmaryrd}
\usepackage{xcolor}

\usepackage{breqn}

\usepackage{hyperref}
\usepackage{algorithm}
\usepackage{algorithmic}
\usepackage[labelfont=bf]{caption}
\usepackage{listings,newtxtt}

\usepackage{etoolbox}
\usepackage{subfiles}
\newtoggle{supplementary}
\toggletrue{supplementary}

\usepackage{nicefrac}       % compact symbols for 1/2, etc.
\usepackage[labelfont=bf]{caption}
\usepackage{shortcuts}

\usepackage[displaymath, mathlines, switch]{lineno}

\title{Spatio-Temporal Alignments: \\
	Optimal transport through space and time}
\author{ Hicham Janati \, | \, Marco Cuturi \, | \, Alexandre Gramfort }
\date{Inria Saclay, France \,| \,  Google AI / CREST - ENSAE, France \, | \, Inria Saclay, France}

\begin{document}

	%\linenumbers
% \pagestyle{headings}



\twocolumn[
	\maketitle
	
%\aistatstitle{Spatio-Temporal Alignments: \\
%Optimal transport through space and time}
]
%
%\author{ Hicham Janati - Marco Cuturi  -Alexandre Gramfort }
% \date{Inria Saclay, France -  Google AI / CREST - ENSAE, France - Inria Saclay, France} ]
%
%

% !TEX root = ../main.tex
\begin{abstract}
    Comparing data defined over space and time is notoriously hard. It involves quantifying both spatial and temporal variability while taking into account the chronological structure of the data. Dynamic Time Warping (DTW) computes a minimal cost alignment between time series that preserves the chronological order but is inherently blind to spatio-temporal shifts. In this paper, we propose Spatio-Temporal Alignments (STA), a new differentiable formulation of DTW that captures spatial and temporal variability. Spatial differences between time samples are captured using regularized Optimal transport. While temporal alignment cost exploits a smooth variant of DTW called soft-DTW. We show how smoothing DTW leads to alignment costs that increase quadratically with time shifts. The costs are expressed using an unbalanced Wasserstein distance to cope with observations that are not probabilities. Experiments on handwritten letters and brain imaging data confirm our theoretical findings and illustrate the effectiveness of STA as a dissimilarity for spatio-temporal data.
\end{abstract}
% !TEX root = ../main.tex
\section{Introduction}
To discriminate between two sets of observations, one must find an appropriate metric that emphasizes their differences. The performance of any machine learning model is thus inherently conditioned by the discriminatory power of the metrics it is built upon. 
Yet, designing the \emph{best} metric for the application at hand is not an easy task. A \emph{good} metric must take into account the structure of its inputs. Here we propose a differentiable metric for spatio-temporal data.

\paragraph{Spatio-temporal data}
Spatio-temporal data consist of time series where each time sample is multivariate and lives in a certain coordinate system equipped with a natural distance. Such a coordinate system can correspond to 2D or 3D positions in space, pixel positions etc. This setting is encountered in several machine learning problems. Multi-target tracking for example, involves the prediction of the time indexed positions of several objects or particles~\citep{doucet02}. In brain imaging, magnetoencephalography (MEG) and functional magnetic resonance imaging (fMRI) yield measurements of neural activity in multiple positions and at multiple time points~\citep{gramfort-etal:2011}.  Quantifying spatio-temporal variability in brain activity can allow to compare different clinial populations.  In traffic dynamics studies, several public datasets report the tracked movements of pedestrians and cars such as the NYC taxi data~\citep{taxi}. 

\paragraph{Optimal transport}
Recently, optimal transport has gained considerable interest from the machine learning and signal processing community~\citep{otbook}. Indeed, when data are endowed with geometrical properties, 
Optimal transport metrics (a.k.a the Wasserstein distance) can capture spatial variations between probability distributions. Given a transport cost function -- commonly referred to as ground metric -- the Wasserstein distance computes the optimal transportation plan between two measures. Its heavy computational cost can be significantly reduced by using entropy regularization \citep{cuturi13}. Besides, when measures are not normalized, it is possible to use the unbalanced optimal transport formulation of \citet{chizat17}, which allows to compute the entropy regularized Wasserstein distance using Sinkhorn's algorithm with minor modifications. To take into account the temporal dimension, one could define the ground metric as a combination of spatial and temporal shifts similarly to the definition of $TL^p$ distances~\citep{thorpe17}. This method however ignores the chronological order of the data and requires a tuning parameter to settle the tradeoff between spatial and temporal transport cost. Instead, one can make use of the dynamic time warping (DTW) framework.

\paragraph{Dynamic time warping}
Given a pairwise distance matrix between all time points of two time series of respective lengths $m, n$, DTW computes the minimum-cost alignment between the time series~\citep{sakoe78} while preserving the chronological order of the data. Indeed, the DTW optimization problem is constrained on alignments where no temporal back steps are allowed. It can be seen as an OT-like problem where the transport plan must not respect the marginal constraints but instead is a binary matrix with at least one non-zero entry per line and per column, and where the cumulated non-zero path is formed by $\rightarrow, \downarrow, \searrow $ steps exclusively. However, the binary nature of this set makes the DTW loss non-differentiable which is a major limitation when DTW is used as a loss function. To circumvent this issue, several authors introduced smoothed versions of DTW \citep{saigo04, cuturi11, cuturi17}. Instead of selecting \emph{the} minimum cost alignment, Global Alignment Kernels (GAK) \cite{saigo04, cuturi11} compute a weighted cost on the whole set of possible alignments. Similarly, the soft-minimum generalization approach of \citet{cuturi17} -- called soft-DTW -- provides a similar framework to that of GAK where gradients can easily be computed used a backpropagation of Bellman's equation~\citep{bellman}.

\paragraph{Our Contributions}
Our contributions are twofold. First, we show that, contrarily to DTW that is blind to time shifts, soft-DTW captures temporal shifts with a quadratic lower bound. Second, we propose to use a divergence based on unbalanced optimal transport as a cost for the soft-DTW loss function. The resulting distance-like function is differentiable and can capture both spatial and temporal differences. We call it Spatio-Temporal Alignment (STA). Since the optimal temporal alignment between two time series is computed by minimizing the overall \emph{spatial} transportation cost, this formulation leads to an intuitive metric to compare time series of spatially defined samples while taking into account the chronological structure of the data.
We experimentally illustrate the relevance of STA on clustering tasks of brain imaging and handwritten letters datasets.

\paragraph{Structure}
Section \ref{s:background} provides some background material on optimal transport and dynamic time warping. We show in section \ref{s:softdtw} that soft-DTW increases at least quadratically with temporal shifts. In Section \ref{s:sta} we introduce the proposed STA dissimilarity. Finally, Section \ref{s:experiments} illustrates the potential applications of STA using several experiments.

\paragraph{Notation}
We denote by $\mathds 1_p$ the vector of ones in $\bbR^p$ and by $\intset{q}$ the set $\{1, \ldots, q\}$ for any integer $q \in \bbN$. The set of vectors in $\bbR^p$ with non-negative (resp. positive) entries is denoted by $ \bbR^p_+$ (resp. $\bbR^p_{++}$).  On matrices, $\log$, $\exp$ and the division operator are applied element-wise. We use $\odot$ for the element-wise multiplication between matrices or vectors. If $\bX$ is a matrix, $\bX_{i.}$ denotes its $i^{\text{th}}$ row and $\bX_{.j}$ its $j^{\text{th}}$ column. We define the Kullback-Leibler (KL) divergence between two positive vectors by $\kl(\bx, \by) = \langle \bx , \log(\bx / \by) \rangle + \langle \by - \bx, \mathds 1_p \rangle$ with the continuous extensions  $0\log(0 / 0) = 0 $ and $0 \log(0) = 0$. We also make the convention $\bx \neq 0 \Rightarrow \kl(\bx | 0) = +\infty$. The entropy of $\bx \in \bbR^n$ is defined as $H(\bx) = - \langle \bx,\log(\bx) - \mathds 1_p \rangle $. The same definition applies for matrices with an element-wise double sum. The feasible set of binary matrices of $\bbR^{m \times n}$ where only $\rightarrow, \downarrow, \searrow$ movements are allowed is denoted by $\cA_{m, n}$.

% !TEX root = ../main.tex

\section{Background on Optimal transport and soft-DTW}
\label{s:background}
\subsection{Unbalanced Optimal transport}

\paragraph{Entropy regularization}
Consider a finite metric space $(E, d)$ where $E = \{1, \dots, p\}$. Let $\bM$ be the matrix where $\bM_{ij}$ corresponds to the distance $d$ between entry $i$ and $j$. Let $\bx, \by$ be two normalized histograms on $E$ ($\bx^\top \mathds 1 = \by^\top \mathds 1 = 1$). Assuming that transporting a fraction of mass $\bP_{ij}$ from $i$ to $j$ is given by $\bP_{ij} \bM_{ij}$, the total cost of transport is given by $\langle \bP, \bM\rangle = \sum_{ij} \bP_{ij} \bM_{ij}$. The Wasserstein distance is defined as the minimum of this total cost with respect to $\bP$ on the polytope $\cP = \{\bP \in \bbR_+^{p\times p}, \bP\mathds 1 = \bx, \bP^\top \mathds 1 = \by\}$~\citep{doklady}. 
Entropy regularization was introduced by \cite{cuturi13} to propose a faster and more robust alternative to the direct resolution of the linear programming problem. Formally, this accounts to minimizing the loss $ \langle \bP, \bM \rangle - \varepsilon H(\bP) $ where $\varepsilon > 0$ is a regularization hyperparameter. Up to a constant, this problem is equivalent to:
\begin{equation}
    \label{eq:ot}
\min_{\bP\in\cP} \ \varepsilon \kl(\bP, e^{- \frac{\bM}{\varepsilon}}) \enspace ,
\end{equation}
which can be solved using Sinkhorn's algorithm.

\paragraph{Unbalanced Wasserstein}
To cope with unbalanced inputs, \citet{chizat17} proposed to relax the marginal constraints of the polytope $\cP$ using a Kullback-Leibler divergence. Given a hyperparameter $\gamma > 0$:
 \begin{equation}
\label{eq:unbalanced-wasserstein}
\begin{aligned}
W(\bx, \by) = \min_{\bP \in {\bbR_+}^{p\times p}} \, &\varepsilon \kl(\bP| e^{- \frac{\bM}{\varepsilon}}) +\\& \gamma \kl(\bP\mathds 1 | \bx) + \gamma \kl(\bP^\top \mathds 1 | \by) \enspace.
\end{aligned}
\end{equation}
While the first term minimizes transport cost, the added Kullback-Leibler divergences penalize for mass discrepancies between the transport plan and the input unnormalized histograms. 
Problem \eqref{eq:unbalanced-wasserstein} can be solved using the following proposition.

\begin{prop}
\label{p:dualw}
Let $\bx, \by \in \bbR^p_+$. The unbalanced Wasserstein distance is obtained from the dual problem:
\begin{equation}
  \begin{aligned}
    \label{eq:dualw}
  W(\bx, \by) = \max_{u, v \in \bbR^p} &- \gamma \langle \bx, e^{-\frac{u}{\gamma}} - 1\rangle - \gamma \langle \by, e^{-\frac{v}{\gamma}} - 1\rangle - \\& \varepsilon \langle e^{\frac{u \oplus v}{\varepsilon}} - 1, e^{-\frac{M}{\varepsilon}}\rangle \enspace .
  \end{aligned}
\end{equation}
Moreover, with the change of variables: 
$\omega = \frac{\gamma}{\gamma + \varepsilon}$, $\bK=e^{-\frac{\bM}{\varepsilon}}, \ba = e^{\frac{u}{\varepsilon}}, \bb = e^{\frac{v}{\varepsilon}}$, the optimal dual points are the solutions of the fixed point problem:
\begin{equation}
    \label{eq:fixedpoint}
      \ba = \left(\frac{\bx}{K\bb}\right)^{\omega} \quad , \quad
    \bb = \left(\frac{\by}{K^{\top}\ba}\right)^{\omega}
\end{equation}

and the optimal transport plan is given by: \begin{equation}
    \label{eq:primaldual}
(\bP_{ij}) = (\ba_i\bK_{ij}\bb_j). 
\end{equation}

\end{prop}

\proof
Since the conjugate of the linear operator $G:\bP \mapsto (\bP\mathds1, \bP^{\top}\mathds 1)$ is given by $G^\star: (u, v) \mapsto u \oplus v$, the Fenchel duality theorem leads to \eqref{eq:unbalanced-wasserstein}. The dual loss function is concave and goes to $-\infty$ when $\|u, v\| \to +\infty$, canceling its gradient yields \eqref{eq:fixedpoint}.
Finally, since the primal problem is convex, strong duality holds and the primal-dual relationship gives \eqref{eq:primaldual}. See \citep{chizat17} for a detailed proof.
\qed

 Solving the fixed point problem \eqref{eq:fixedpoint} is equivalent to alternate maximization of the dual function \eqref{eq:dualw}. Starting from two vectors $\ba, \bb$ set to $\mathds 1$, the algorithm iterates through the scaling operations \eqref{eq:fixedpoint}. This is a generalization of the Sinkhorn algorithm which corresponds to  $\omega = 1$ or $\gamma = +\infty$.
\begin{corollary}
	\label{cor:symmetricw}
	Let $\bx \in \bbR^p_+$. The associated optimal dual scalings $\ba$, $\bb$ to computing $W(\bx, \bx)$ are given by the solution of the fixed point problem:
	$ \bb = \ba = \left(\frac{\by}{K\ba}\right)^{\phi} $
\end{corollary}
\proof The symmetry of the dual problem \eqref{eq:dualw} with $\bx = \by$ implies immediately that $\ba = \bb$. Proposition \ref{p:dualw} gives the fixed point equation.
\subsection{Soft Dynamic Time Warping}
\paragraph{Forward recursion}
Let $\bx = (\bx_1^\top, \dots, \bx_{T_1}^\top)^\top \in \bbR^{T_1, p}$ and $\by = (\by_1^\top, \dots, \by_{T_2}^\top)^\top \in \bbR^{T_2, p}$ be two time series of respective lengths $T_1, T_2$ and dimension $p$. The set of all feasible alignments in a $(T_1, T_2)$ rectangle is denoted by $\cA_{T_1, T_2}$. Given a pairwise distance matrix $\Delta(\bx, \by) \eqdef (\delta(\bx_i, \by_j))_{ij}$, soft-DTW is defined as:
\begin{equation}
    \label{eq:sdtw}
    \sdtw(\bx, \by; \Delta) = {\softmin}_{\beta}\{\langle \bA, \Delta(\bx, \by)\rangle, \bA \in \cA_{T_1, T_2}\} \enspace ,
\end{equation}
where the soft-minimum operator of a set $\cA$ with parameter $\beta$ is defined as:
\begin{equation}
    {\softmin}_{\beta}(\cA) =
    \left\{\begin{array}{ll}
        -\beta \log\left(\sum_{\cA} e^{- \nicefrac{a}{\beta}}\right) \text{\quad if } \beta > 0\\
        \min_{a \in \cA} a \text{\quad if } \beta = 0
        \end{array} \right.
\end{equation}

\begin{figure}[t]
    \centering
	\includegraphics[width=0.7\linewidth, trim={1.cm 0 1.cm 0},clip]{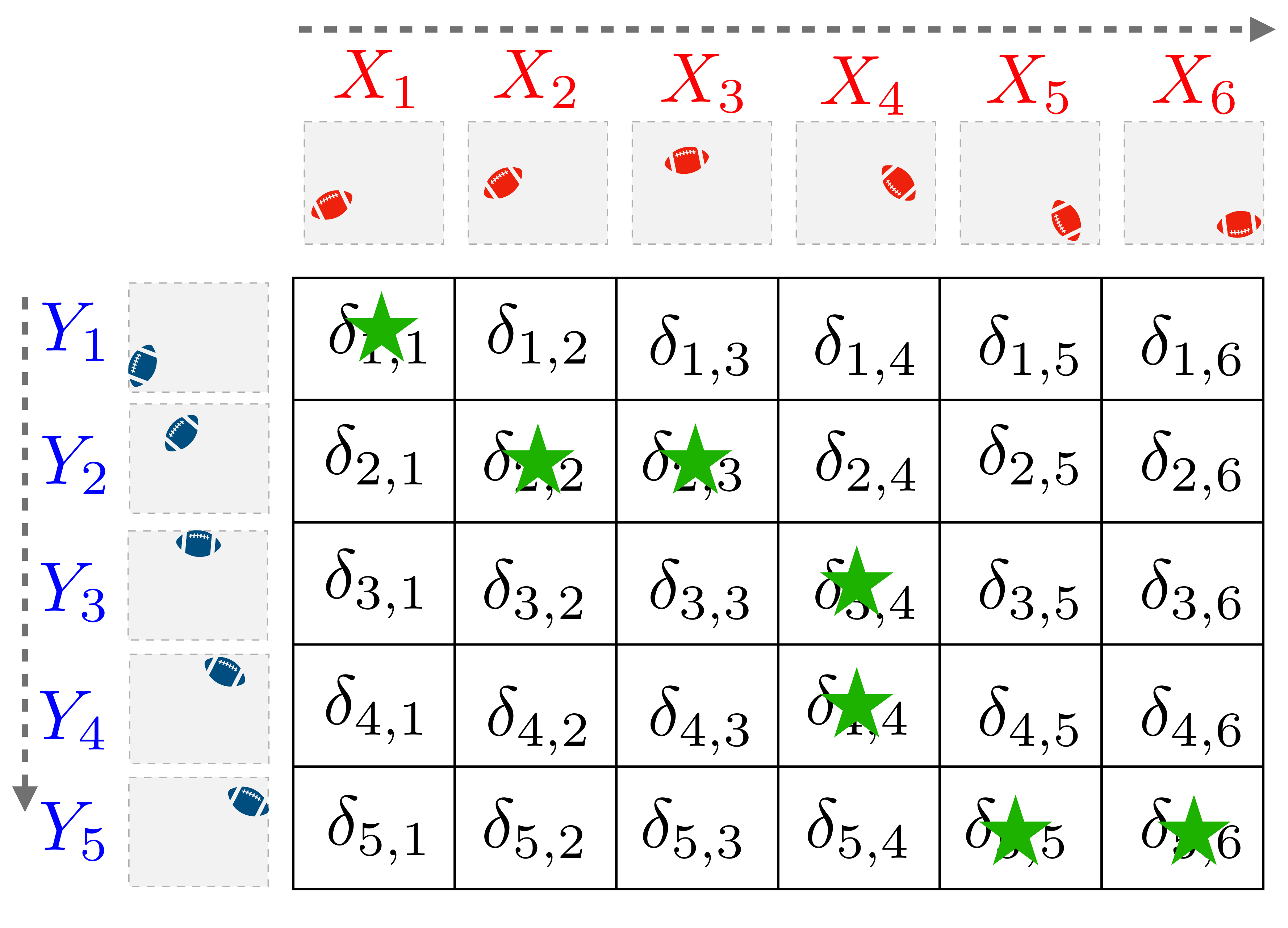}
	\caption{Example of Dynamic time warping alignment between two time series of images given a pairwise distance matrix. \label{f:dtw-example}}
\end{figure}

Figure~\ref{f:dtw-example} illustrates two time series of images and their cost matrix $\Delta$. The path from (1, 1) to (5, 6) is an example of a feasible alignement in $\cA_{5, 6}$.
When $\beta = 0$, the soft-minimum is a minimum and $\sdtw$ falls back to the classical DTW metric. Nevertheless, it can still be computed using the dynamic program of Algorithm \ref{a:dynamicprogram} with a soft-min instead of min operator. % It is important to notice that Algorithm~\ref{a:dynamicprogram} stores the whole intermediary alignment matrix $R$ which is needed for automatic differentiation.

\begin{algorithm}[t]
   \caption{BP recursion to compute $\sdtw$ \citep{cuturi17} \label{a:dynamicprogram}}
\begin{algorithmic}
   \STATE {\bfseries Input:}  data $\bx, \by$ soft-min parameter $\beta$ and distance function $\delta$ 
      \STATE {\bfseries Output:} $\sdtw(\bx, \by) = r_{T_1, T_2}$
    \STATE $r_{0,0} = 0; r_{0,j} = r_{i, 0} =\infty$ for $i \in \intset{T_1}$, $j \in \intset{T_2}$ 
    \FOR{$i=1$ {\bfseries to} $T_1$}
    \FOR{$j=1$ {\bfseries to} $T_2$}
    \STATE $r_{i,j} = \delta(\bx_i, \by_j)$ ${\softmin}_\beta(r_{i-1, j-1}, r_{i-1, j}, r_{i, j-1})$
      \ENDFOR
     \ENDFOR
\end{algorithmic}
\end{algorithm}

\section{Soft-DTW captures time shifts}
\paragraph{Temporal shifts}
\label{s:softdtw}
Let $\bx$ and $\by$ be two time series. When studying the properties of $\sdtw$, the dimensionality of the time series is irrelevant since it is compressed when computing the cost matrix $\Delta$. Thus, to study temporal shifts, we assume in this section that $\bx$ and $\by$ are univariate and belong to $\bbR^{T}$. To properly define temporal shifts, we introduce a few preliminary notions.  We name the first (respectively, last) time index where $\bx$ fluctuates the \emph{onset} (respectively, the \emph{offset}) of $\bx$ and denote it by $\on(\bx)$ (respectively, $\off(\bx)$). The \emph{fluctuation set}  of $\bx$ is denoted by $\fluc(\bx)$ and corresponds to all time indices between the onset and the offset. Formally:
\begin{align}
    \on(\bx) &= \argmin_{i \in \intset{1, T - 1}} \{\bx_{i+1} \neq \bx_i\}     \label{eq:onset} \\
    \off(\bx) &= \argmax_{i \in \intset{1, T - 1}} \{\bx_{i+1} \neq \bx_i\}     \label{eq:offset} \\
	\fluc(\bx) & = \{ i \in \intset{1, T}, \on(\bx) \leq i \leq \off(\bx)\} \label{eq:fluctuation}
\end{align}
For $\bx$ and $\by$ to be temporally shifted with respect to each other,  their values must agree both within and outside their (different) fluctuation sets.
\begin{definition}[Temporal k-shift]
\label{def:temporalshift}
Let $\bx$ and $\by$ be two time series in $\bbR^T$ and $k \in \intset{1, T-1}$. We say that $\by$ is temporally k-shifted with respect to $\bx$ and write $\by = \bx_{+_k}$ if and only if:
\begin{equation}
\begin{aligned}
	&\on(\by) = \on(\bx) + k \\
	& \off(\by)  = \off(\bx) + k \\
	& i \leq \on(\bx),  j \leq \on(\by) \Rightarrow \bx_i = \by_j \\
   &  i \geq \off(\bx),  j \geq \off(\by) \Rightarrow \bx_i = \by_j  \\
   & i \in \fluc(\bx), j \in \fluc(\by), |i - j| = k \Rightarrow \bx_i = \by_j \enspace .
\end{aligned}
\end{equation}
\end{definition}
\begin{figure}[t]
	\includegraphics[width=\linewidth]{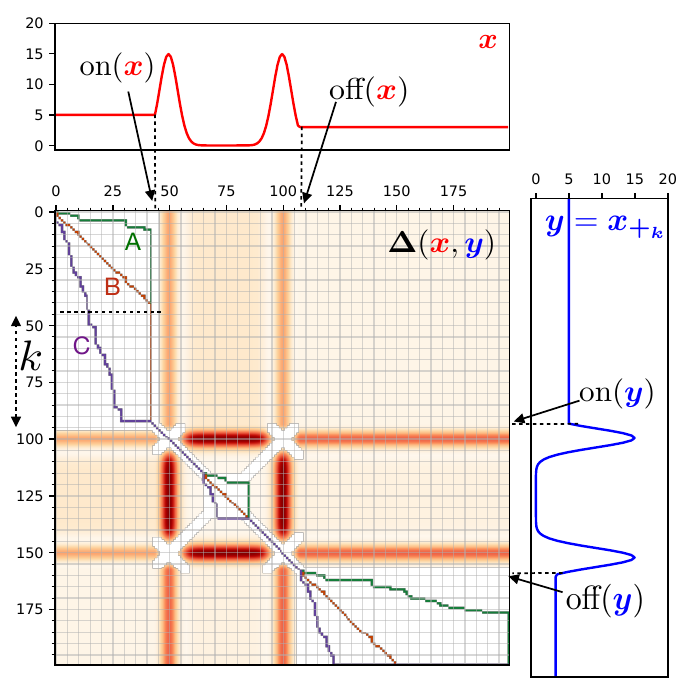}
	\caption{Example of 3 DTW alignment paths (A, B and C) between $\bx$ and $
	\by = \bx_{+_k}$ with a temporal 50-shift. The heatmap of the distance matrix $\Delta$ shows (white) rectangles where all paths A, B, C have an equal DTW cost of 0. These areas correspond to time durations where $\bx$ and $\bx_{+_k}$ are constant. It is noteworthy that when shifting one time series, among the areas crossed by the alignments A, B, C, only the two white rectangles outside the fluctuation set change in size. \label{f:shiftexample}}
\end{figure}
An example of a temporal 50-shift is illustrated in Figure~\ref{f:shiftexample}. The heatmap of the squared Euclidean cost matrix $\Delta$ shows three rectangular white areas where all alignments A, B and C have the same cost of 0. Since $\dtw$ is defined as the minimum of all alignment costs, all these paths are equivalent. Temporal k-shifts change the set of alignments with cost 0 but do not change the $\dtw$ value. However, when $\beta>0$, $\sdtw$ computes a weighted sum of all possible paths, which is affected by temporal shifts by including the number of equivalent paths. The cardinality of $\cA_{m, n}$ is known as the Delannoy number $D(m - 1, n - 1)$~\citep{cuturi11}.  For the sake of convenience, we consider the shifted Delannoy sequence starting at $n=m=1$ so that: $\card(\cA_{m, n}) = D_{m, n}$. If $\beta$ is positive but small enough, the alignements with 0 cost dominate the $\sdtw$ logsumexp. This leads to proposition~\ref{prop:kshift}.
\begin{definition}[Delannoy sequence]
		\label{def:delannoy}
	The Delannoy number $D_{m, n}$ corresponds to the number of paths from $(1, 1)$ to $(m, n)$ in a $(m \times n)$ lattice where only  $\rightarrow, \downarrow, \searrow$ movements are allowed. It can also be defined with the recursion $\forall m, n \in \bbN^{\star}$:
	\begin{align}
		& D_{1, n} = D_{m, 1} = 1 \\
		&  D_{m + 1, n + 1} = D_{m , n + 1} + D_{m + 1, n} + D_{m , n }  \label{eq:delannoy-rec} \enspace .
	\end{align}
\end{definition}
\begin{prop}
	\label{prop:kshift}
	 Let $k \in \intset{1, T - 1}$, let $m = \on(\bx)$ and $m' = T - \off(\bx)$. Let $\mu = \min_{i,j}\{\Delta(\bx, \bx)_{ij} | \Delta(\bx, \bx)_{ij} > 0\}$. If   $0 < \beta \leq \frac{\mu}{\log(3T D_{T, T})}$ :
\begin{align}
\label{eq:kshift}
\sdtw(\bx, \bx_{+_k}) -  \sdtw(\bx, \bx)  \geq \nonumber \\
\beta \log \left( \frac{ D_{m, m}  D_{m' , m'} }{ D_{m + k, m}  D_{m' - k , m'} } \right) - \frac{\beta}{3T}
\end{align}
\end{prop}
\proofsketch
When $\beta$ is small, the logsumexp in the $\sdtw$ is dominated by the number of alignments with 0 cost. This number is given by:
$D_{\on(\bx) , \on(\by) } \Omega  D_{T - \off(\bx) , T - \off(\by)}$, where $\Omega$ is the number of 0 cost alignments within the cross product of the fluctuation sets. However, temporal shifts do not change $\Omega$ but only change the outermost sets. For instance, considering the example of Figure~\ref{f:shiftexample} one can see that only rectangles outside the fluctuation set are affected. Therefore, $\Omega$ cancels out in the first term of \eqref{eq:kshift}.
Using the upper bound on $\beta$, we derive the second term. The full proof is provided in the supplementary materials. \qed
\paragraph{Quadratic lower bound}
The purpose of the rest of the section is to find a lower bound of the right side of \eqref{eq:kshift} that depends on $k$. To do so, we incrementally upper bound the off-diagonal Delannoy number $D_{m, m+k}$ with its left and bottom neighbors. When $k=1$,  the following Lemma happens to be crucial to derive the lower bound.
\begin{lemma}[Bounded growth]
	\label{lem:growth}
	Let $c = 1 + \sqrt{2}$ and $m \geq 1$. The  central (diagonal) Delannoy numbers $D_m = D_{m, m}$ verify:
	\begin{equation}
	\label{eq:growth}
	D_{m+1} \leq c^2 D_{m}
	\end{equation}
\end{lemma}
\proof The proof is provided in the supplementary materials.
\begin{prop}
	\label{prop:delannoy-ineq}
	Let $c = 1 + \sqrt{2}$.  $\forall m, i \in \bbN^\star$:
	\begin{align}
	\label{eq:delannoy-ineq}
	 &D_{m, m + i}  \leq c \Phi_{m, i} D_{m, m + i - 1} \\ 
 	 &c \Psi_{m, i} D_{m, m + i}  \leq D_{m + 1, m + i}
	\end{align}
	Where
	\[
	\left\{	 \begin{array}{l} 
					       \Phi_{m, i} = 1 - \frac{ (1 - \frac{1}{c}) (i - 1) + \frac{1}{c}}{m + i - 1} \\
					        \Psi_{m, i} = 1 + \frac{(1 - \frac{1}{c}) (i - 1)}{m} 
					        \end{array}
				    \right.
	\]
\end{prop}
\proofsketch We prove both statements jointly with a double recurrence reasoning. The initializing for $i=1$ is immediately obtained using the bounded growth Lemma~\ref{lem:growth}. 
To show the induction step, we rely on the recursion equation \eqref{eq:delannoy-rec}. For the sake of brevity, the full proof is provided in the supplementary materials.

By applying proposition \ref{prop:delannoy-ineq} to all $i \in \intset{1, k}$, the product of all the obtained inequalities leads to a bound on the right side of proposition \ref{prop:kshift}:
\begin{prop}
	\label{prop:bound}
		 Let $k \in \intset{1, T - 1}$, let $m = \on(\bx)$ and $m' = T - \off(\bx)$. Using the notations of proposition \ref{prop:delannoy-ineq} for $\Phi$ and $\Psi$:
	\begin{equation}
	\label{eq:bound}
		\log \left( \frac{ D_{m, m}  D_{m' , m'} }{ D_{m + k, m}  D_{m' - k , m'} } \right) \geq  \log\left( \prod_{i=1}^k \frac{\Psi_{m' - i, i}}{\Phi_{m, i}}  \right) 
	\end{equation}
\end{prop}
\proof
Iterating the inequalities of proposition ~\ref{prop:delannoy-ineq}, we have on one hand with the first inequality:
\begin{equation}
    \label{eq:proof-ineq1}
    \frac{ D_{m, m}}{D_{m , m + k}}  \geq \frac{1}{c^k \prod_{i=1}^{k} \Phi_{m, i} } \enspace ,
\end{equation}
and on the other hand with the second inequality:
$$
    \frac{D_{m + k, m + k} }{ D_{m , m + k} } \geq  c^k \prod_{i=0}^{k-1} \Psi_{m + i, k - i} = c^k \prod_{i=1}^{k} \Psi_{m + k - i, i}  \enspace .
$$
With the change of variable $m' = m + k$ and the symmetry of Delannoy numbers, we have:
\begin{equation}
\label{eq:proof-ineq2}
\frac{D_{m', m'} }{ D_{m' , m' - k} } \geq  c^k \prod_{i=1}^{k} \Psi_{m' - i, i} \enspace .
\end{equation}
Taking the product of \eqref{eq:proof-ineq1} and \eqref{eq:proof-ineq2} and the result of proposition~\ref{prop:kshift} concludes the proof. \qed

Finally, we can now state our main theorem.
\begin{theorem}
	 Let $k \in \intset{1, T - 1}$, let $m = \on(\bx)$ and $m' = T - \off(\bx)$. Let $\mu = \min_{i,j}\{\Delta(\bx, \bx)_{ij} | \Delta(\bx, \bx)_{ij} > 0\}$. If   $0 < \beta \leq \frac{\mu}{\log(3T D_{T, T})}$ :
	\label{thm:bound}
		\begin{equation}
	\label{eq:quadratic}
	\sdtw(\bx, \bx_{+_k}) -  \sdtw(\bx, \bx) \geq \beta \alpha k(k-1) +  \beta \rho k
	\end{equation}
	Where $\alpha = \frac{2 - \sqrt{2}}{2}(\frac{1}{m'} + \frac{1}{m + m'}) > 0$ and $\rho = \frac{3\sqrt{2} -4 }{3T} > 0$.
\end{theorem}
\proof
Let $a = 1 - \frac{1}{c}$.
Developping the bound in proposition \ref{prop:bound}, we get:
\begin{equation}
\label{eq:bound-log} 
\log\left( \prod_{i=1}^k \frac{\Psi_{m' - i, i}}{\Phi_{m, i}}  \right)  =  \sum_{i=1}^k \log(\Psi_{m' - i, i}) - \log(\Phi_{m, i})
\end{equation}
Using the inequality $ \frac{x}{1 + x} \leq \log(1 + x) \leq  x$ for  $x > -1$ on both logarithms we have, on one hand:
\begin{align}
\label{eq:logpsi}
 \log(\Psi_{m' - i, i}) &= \log\left( 1 + \frac{a(i - 1)}{m' - i} \right)\nonumber \\
 	 						        & \geq \frac{a(i - 1)}{m' - i + a(i-1)} = \frac{a(i-1)}{m' - \frac{i}{c}-a} \nonumber  \\
 	 						        & \geq \frac{a(i-1)}{m'}
\end{align}
and on the other hand:
\begin{align}
\label{eq:logphi}
- \log(\Phi_{m, i}) &= -  \log\left( 1 - \frac{ a (i - 1) + \frac{1}{c}}{m + i - 1}  \right) \nonumber \\
							    & \geq \frac{a(i - 1) + \frac{1}{c}}{m + i - 1} \geq \frac{a(i - 1) + \frac{1}{c}}{m + m'} \nonumber \\
	                &\geq\frac{a(i-1)}{m + m'} + \frac{1}{cT} \enspace .
\end{align}
Finally, combining equations \eqref{eq:logpsi} and \eqref{eq:logphi}, the formula $\sum_{i = 1}^k (i-1) = \frac{k(k - 1)}{2}$ and adding the term $-\frac{\beta}{3T}$ of \eqref{eq:kshift} leads the desired quadratic function.  \qed

\begin{figure}[t]
	\centering
	\includegraphics[width=0.8\linewidth]{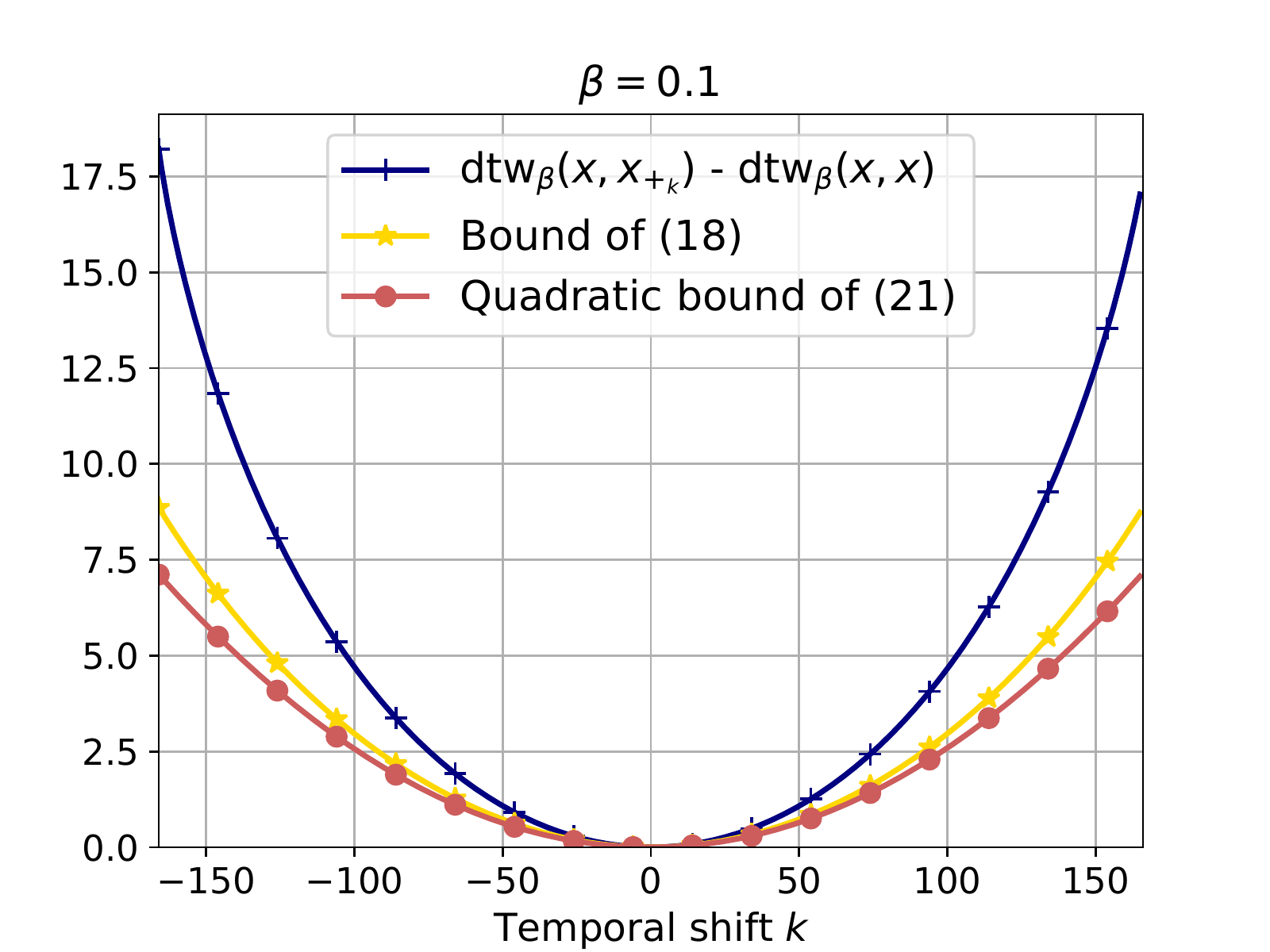}
	\caption{Illustration of the bounds of proposition~\ref{prop:bound} and theorem~\ref{thm:bound} with $\beta=0.1$ and $T=400$. The time series $\bx$ is a centered version of the one displayed in Figure~\ref{f:shiftexample}.  \label{f:bound}}
\end{figure}

We illustrate these bounds experimentally with the example of Figure~\ref{f:shiftexample} with $T = 400$ to allow for larger temporal shifts and $\beta = 0.1$. Figure~\ref{f:bound} shows that $\sdtw$ is indeed polynomial in $k$; the quadratic bound is sufficient as an approximation for the result of theorem~\ref{thm:bound}. Experimentally, we notice that the assumption on $\beta$ is too restrictive. Indeed, the comparison empirically holds for larger values of $\beta$ which may be desirable in practice to capture more temporal differences. The corresponding figures are provided in the appendix.

% !TEX root = ../main.tex
\section{Spatio-Temporal Alignments}
\label{s:sta}

\paragraph{Unbalanced Sinkhorn divergence}
To capture spatial variability, we propose to use a cost function based on the unbalanced Wasserstein distance~$W$. Since $W(\bx, \bx) \neq 0 $, the resulting metric would fail to identify identical samples. Similarly to the introduction of Sinkhorn divergences for the balanced case \citep{genevay18}, we define the unbalanced Sinkhorn divergence $S$ between two histograms in $\bbR^p_+$ as:
\begin{equation}
    \label{eq:sdivergence}
    S(\bx, \by) = W(\bx, \by) - \frac{1}{2}\left(W(\bx, \bx) + W(\by, \by)\right)
\end{equation}
The proposed dissimilarity -- \emph{Spatio-Temporal Alignement} -- corresponds to the soft-DTW loss with the divergence $S$ as an alignement cost:
\begin{definition}[STA]
\label{def:wdtw}
We define the STA loss as:
\begin{equation}
    \label{eq:wdtw}
    \sta(\bx, \by) = \sdtw(\bx, \by; S)
\end{equation}
\end{definition}
Aside from $S(\bx,~\bx)=~0$, we do not know much about $S$. The rest of this section aims at showing some of its useful properties when $\bK$ is positive semi-definite: non-negativity and coercivity. The curvature of $S$ is however harder to analyze. Nonetheless, it is minimized at $\bx = \by$ and if $\bK$ is, the only stationary points are $\bx = \by$.

\paragraph{Non-negativity}
We show that $S$ is non-negative, we assume that the kernel $\bK = e^{-\frac{\bM}{\varepsilon}}$ is positive semi-definite. This is the case for example with $\bM_{ij} = \|m_i - m_j\|^l$ with $ 0 < l \leq 2$ if the support of the measures is given by $\{m_1, \dots, m_p \in \bbR\}$.

\begin{prop}
	\label{prop:nonnegativeS}
	Let $\bx, \by \in \bbR_+^{p}$. If $\bK = e^{-\frac{ \bM}{\varepsilon}}$ is positive semi-definite:
	$$ S(\bx, \by) \geq 0 $$
	Moreover, if $\bK$ is positive definite, $S(\bx, \by) = 0 \Rightarrow \bx = \by$.
\end{prop}
\proof
Let $\bc$ and $\bd$ denote the solutions of the fixed point problems:
$\bc = \left(\frac{\bx}{K\bc}\right)^{\omega} $ and $\bd = \left(\frac{\by}{K\bd}\right)^{\omega} $. With the change of variable $\ba = e^{-\frac{\bu}{\varepsilon}} $ and $\bb = e^{-\frac{\bv}{\varepsilon}}$, let $(\ba, \bb) \to \cD(\ba, \bb)$ denote the dual function of \eqref{eq:dualw}.  On one hand, by Corollary \ref{cor:symmetricw},
$W(\bx, \bx) = \max_{\ba, \bb \in \bbR_+^{p}} \cD(\ba, \bb) = \cD(\bc, \bc).$ Similarly, $W(\by, \by) = D(\bd, \bd)$. On the other hand, by definition of the max  $W(\bx, \by) \geq \cD(\bc, \bd)$. Therefore:
\begin{align*}
 S(\bx, \by) &\geq D(\bc, \bd) - \frac{1}{2}(D(\bc, \bc) +  D(\bd, \bd)) \\
 	&= \varepsilon \left[ - \langle \bc \otimes \bd, K \rangle +\frac{1}{2} \langle \bc \otimes \bc, K \rangle + \frac{1}{2}\langle \bd \otimes \bd, K \rangle\right] \\
 	&= \varepsilon \left[ - \langle \bc , K\bd \rangle +\frac{1}{2} \langle \bc,  K\bc \rangle + \frac{1}{2}\langle \bd , K\bd \rangle\right] \\
 	&= \frac{\varepsilon}{2}\langle \bc - \bd, \bK(\bc - \bd)\rangle \geq 0
\end{align*}
Where the last inequality follows from the positivity of $\bK$. If $\bK$ is positive definite, the last inequality is strict unless $\bc = \bd$, in which case the fixed point equations lead to $\bx =\by$. \qed

\paragraph{Coercivity}
Regardless of the nature of $\bK$,  we will now show that $S(., \by)$ is coercive for any fixed~$\by$. To do so, we first show that $S$ only depends on the sums of transported mass:

\begin{prop}
	\label{prop:Smass}
		Let $\bx, \by \in \bbR_{+}^{p}$ and $\bP_{\bx, \by} \in \bbR_+^{p\times p}$ their associated transport plan, solution of \eqref{eq:unbalanced-wasserstein}. Then:
		\begin{equation}
		\label{eq:Smass}
		S(\bx, \by) = (\varepsilon + 2 \gamma) (\frac{1}{2} \|P_{\bx, \bx}\|_1 + \frac{1}{2} \|P_{\by, \by}\|_1 -  \|P_{\bx, \by}\|_1 )
		\end{equation}
\end{prop}
\proofsketch
Let $\bx, \by \in \bbR^{p}_+$. And let $\ba, \bb$ the dual scalings associated with the dual problem of $W(\bx, \by)$. The corresponding primal solution is given by $\bP_{ij} = \ba_i \bK_{ij}\bb_j$. Therefore, using the fixed point equations \eqref{eq:fixedpoint}, we have: $\|\bP_{\bx, \by}\|_1 = \langle \ba, \bK\bb\rangle = \langle \bx, \ba^{-\frac{\varepsilon}{\gamma}}\rangle = \langle \bb, \bK^{\top}\ba\rangle = \langle \by, \bb^{-\frac{\varepsilon}{\gamma}}\rangle$. Therefore, at optimality, the dual function \eqref{eq:dualw} is equal to:
$$W(\bx, \by) = - (\varepsilon + 2 \gamma) \|\bP_{\bx, \by}\|_1 + \gamma (\|\bx\|_1 + \|\by\|_1) \enspace + \varepsilon \|\bK\|_1,
$$
Writing $W(\bx, \bx)$ and $W(\by, \by)$ in the same way ends the proof.
\qed

To prove that $S$ is coercive, we bound $\|\bP_{\bx, \by}\|_1$ with the $\ell_1$ norms of $\bx$ and $\by$:

\begin{lemma}
	\label{lem:planmass}
	Let $\bx, \by \in \bbR_{+}^{p}$ and $\bP_{\bx, \by} \in \bbR_+^{p\times p}$ their associated transport plan, solution of \eqref{eq:unbalanced-wasserstein}. Let $\kappa~=~\min_{i, j} e^{-\frac{M_{ij}}{\gamma}}$. We have the following bounds on the total transported mass:
	\begin{equation}
	\label{eq:planmass}
	\kappa \|\bx\|_1 \|\by\|_1 \leq \| \bP_{\bx, \by}\|^{2 + \frac{\varepsilon}{\gamma}}_1 \leq p^{\frac{3}{2}} \|\bx\|_1 \|\by\|_1
	\end{equation}
\end{lemma}
\proofsketch Writing the first order optimality condition of \eqref{eq:unbalanced-wasserstein} links the optimal transport plan $\bP$ with the inputs $\bx, \by$. The bounds can be easily derived using basic inequalities. For the sake of brevity, the full proof is provided in the supplementary materials.
\begin{prop}
\label{prop:coercive}
For $\by \in \bbR^{p}_+$, the function $\bx \mapsto S(\bx, \by)$ is coercive.
\end{prop}
\proof
 Lemma \ref{lem:planmass} and proposition \ref{prop:Smass}, we get, with $\zeta = \frac{1}{2 + \frac{\varepsilon}{\gamma}}$:
\begin{equation}
\label{eq:Sbound}
 S(\bx, \by)  \geq \kappa (\|\bx\|_1^{2\zeta} + \|\by\|_1^{2\zeta}) - p^{\frac{3}{2}} \|\bx\|_1^{\zeta} \|\by\|_1^{\zeta} 
\end{equation}
Therefore:
$\|\bx\|_1 \to +\infty \Rightarrow S(\bx, \by)~\to~+\infty$ \qed
\paragraph{Differentiability}
$W(., \by)$ is differentiable, and its gradient is given by $\gamma(1 - \ba^{-\frac{\varepsilon}{\gamma}})$ where $\ba$ is the solution of the fixed equation \eqref{eq:fixedpoint} \citep{feydy17}. Thus, $S$ is also differentiable. If $\bK$ is positive semi-definite then $S \geq 0$ and thus, from the following proposition we conclude that all its stationary points are minimizers:
\begin{prop}
	\label{prop:stationarypoints}
	Let $\by, \bx  \in \bbR^{n\times p}_{++}$ be a stationary point of $S$ i.e $\nabla S(\bx, \by) = (\boldsymbol{0}, \boldsymbol{0})$. Then,
	$S(\bx, \by) = 0$. 
	Moreover, if $\bK$ is positive definite, then $\bx = \by$.
\end{prop}
\proof the proof is provided in the appendix.

\paragraph{Optimal transport hyperparameters}
The unbalanced $W$ metric is defined by two hyperparameters: $\varepsilon$ and $\gamma$. On one hand, higher values of $\varepsilon$ increase the curvature of the minimized loss function thereby accelerating the convergence of Sinkhorn's algorithm. This gain in speed is however at the expense of entropy blurring of the transport plan. On the other hand, $\varepsilon \to 0$ leads to a well documented numerical instability that can be mitigated using log-domain stabilization \citep{otbook}. Here we set $\varepsilon$ to the lowest stable value. A practical scale is provided by taking values of $\varepsilon$ proportional to $\frac{m}{p}$ where $m$ is the median of the ground metric $\bM$. The marginals parameter $\gamma$ must be large enough to guarantee transportation of mass. When $\gamma \to 0$, the optimal transport plan $\bP^\star \to \bK$. Large $\gamma$ however slows down the convergence of Sinkhorn's algorithm, especially if the input histograms have significantly different total masses. We set $\gamma$ at the largest value guaranteeing a minimal transport mass using the heuristic proposed in \citep{janati19}.

\paragraph{Complexity analysis}
As shown by Algorithm \ref{a:dynamicprogram}, soft-DTW is quadratic in time. Computing the Sinkhorn divergence matrix is quadratic in $p$. Moreover, when the time series are defined on regular grids such as images, one could benefit from spatial Kernel separability as introduced in \citep{solomon15}. This trick allows to reduce the complexity of Sinkhorn on 2D data from $O(p^2)$ to $O(p^{\frac{3}{2}})$. Moreover, to leverage fast matrix products on GPUs, computing each of the matrices $(W(\bx_i, \by_j))_{ij}$ (resp. $ (W(\bx_i, \bx_i))_{i}, (W(\by_j, \by_j))_{j}$) can be done in a $(n\times m)$ parallel version of Sinkhorn's algorithm, where each kernel convolution $\bK\bv$, $\bK^\top\bu$ is applied to all $nm$ (resp. $m$, $n$) dual variables at once.

\paragraph{Signed data}
The divergence $S$ is defined for non-negative signals only which can be encountered in practice as non-normalized intensities. Yet, one can easily extend $\sta$ to signed data by talking the absolute values of the signals. Or computing $S$ on positive and negative parts separately before averaging.

% !TEX root = ../main.tex

\section{Experiments}
\label{s:experiments}
Our main theoretical result states that $\sdtw$ captures temporal shifts only if $\beta > 0$. Moreover, with the unbalanced Wasserstein divergence as a cost, our proposed dissimilarity $\sta$ should capture both spatial and temporal variability. We illustrate this in a brain imaging simulation and a clustering problem of handritten letters.

\subsection{Brain imaging}
\begin{figure}[t]
\includegraphics[width=0.9\linewidth]{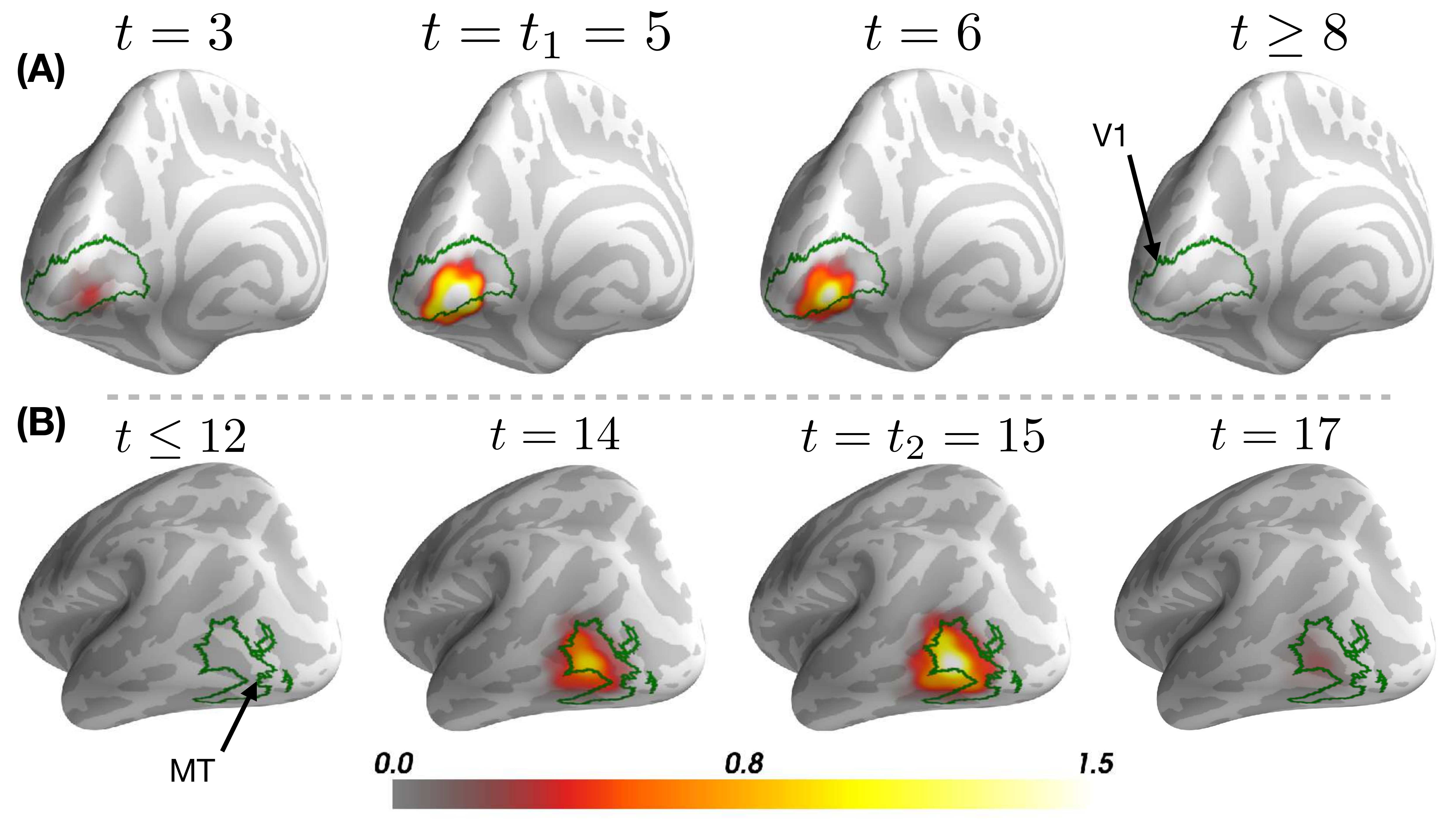}
\caption{Two examples of the simulated time series. \textbf{(A)} brain signal in V1 with a peak at $t=t_1$. \textbf{(B)} brain signal in MT with a peak at $t=t_2$. The borders of the brain regions V1 and MT are highlighted in green. \label{f:brains}}
\end{figure}

\begin{figure}[t]
\centering
\includegraphics[width=0.75\linewidth]{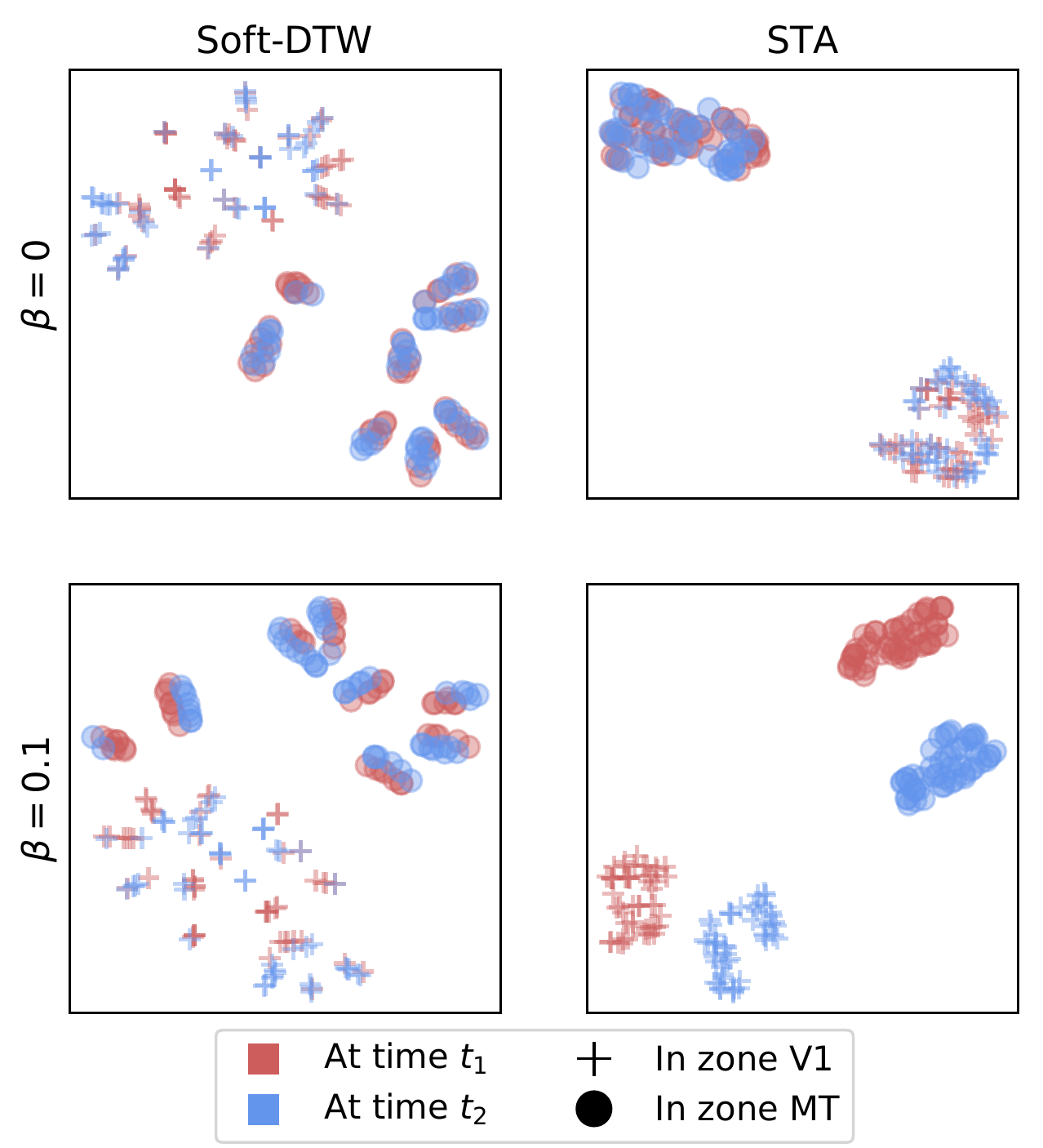}
\caption{t-SNE visualization of the simulated brain signals in two different regions, at two different time instants. With $\beta > 0$, $\sta$ can discriminate between all four groups. \label{f:tsne-meg}}
\end{figure}

Brain imaging data recordings report the brain activity both in space and time. Thanks to their high temporal resolution, Electroencephalography and Magnetoencephalography can capture response latencies in the order of a millisecond. Abnormal differences in latency, amplitude and topography of brain signals are important biomarkers of several conditions of the central nervous system such as multiple sclerosis \citep{whelan10} or amblyopia~\citep{sokoi}. 
We argue here that $\sta$ can aggregate all these differences in a meaningful dissimilarity score. To illustrate this, we use the average brain surface derived from real MRI scans and provided by the FreeSurfer software. We compute a triangulated mesh of 642 vertices on the left hemisphere and simulate 4 types of signals as follows. We set $T=20$ and select 2 activation time points $t_1=5$ and $t_2=15$. We also select two brain regions in the visual cortex given by FreeSurfer's segmentation known as \emph{V1} (primary visual cortex) and \emph{MT} (middle temporal visual area) which are defined on 17 and 8 vertices respectively. Each generated time series peaks at $t_1$ or $t_2$, in a random vertex in V1 or MT with a random amplitude between 1 and 3. For the signals to be more realistic, we apply a Gaussian filter along the temporal and the spatial axes. Examples of the generated data are displayed in Figure~\ref{f:brains}.
We generate $N=200$ samples (50 per time point / brain region) and compute the pairwise dissimilarity matrices $\sdtw$ and $\sta$ with $\beta =0$ and $\beta = 0.1$. Figure~\ref{f:tsne-meg} shows the t-distributed Stochastic Neighbor Embedding (t-SNE) \citep{maaten08, scikit-learn} of the data. As expected, $\sdtw$ cannot capture spatial variability regardless of $\beta$. With $\beta = 0$, $\sta$ separates the data according to the brain region only. Only with positive $\beta$ can $\sta$ identify all four groups. Computing the full $\sta$ dissimilarity matrix required performing $\frac{1}{2} N (N+1) \times T^2 = 8040000$ Sinkhorn loops between 642 dimensional inputs. The whole experiment completed in 10 minutes on our DGX-1 station. Python code and data can be found in \href{https://github.com/hichamjanati/spatio-temporal-alignements}{https://github.com/hichamjanati/spatio-temporal-alignements}.

\subsection{Handwritten letters}

\begin{figure}[t]
    \centering
    \includegraphics[width=0.8\linewidth]{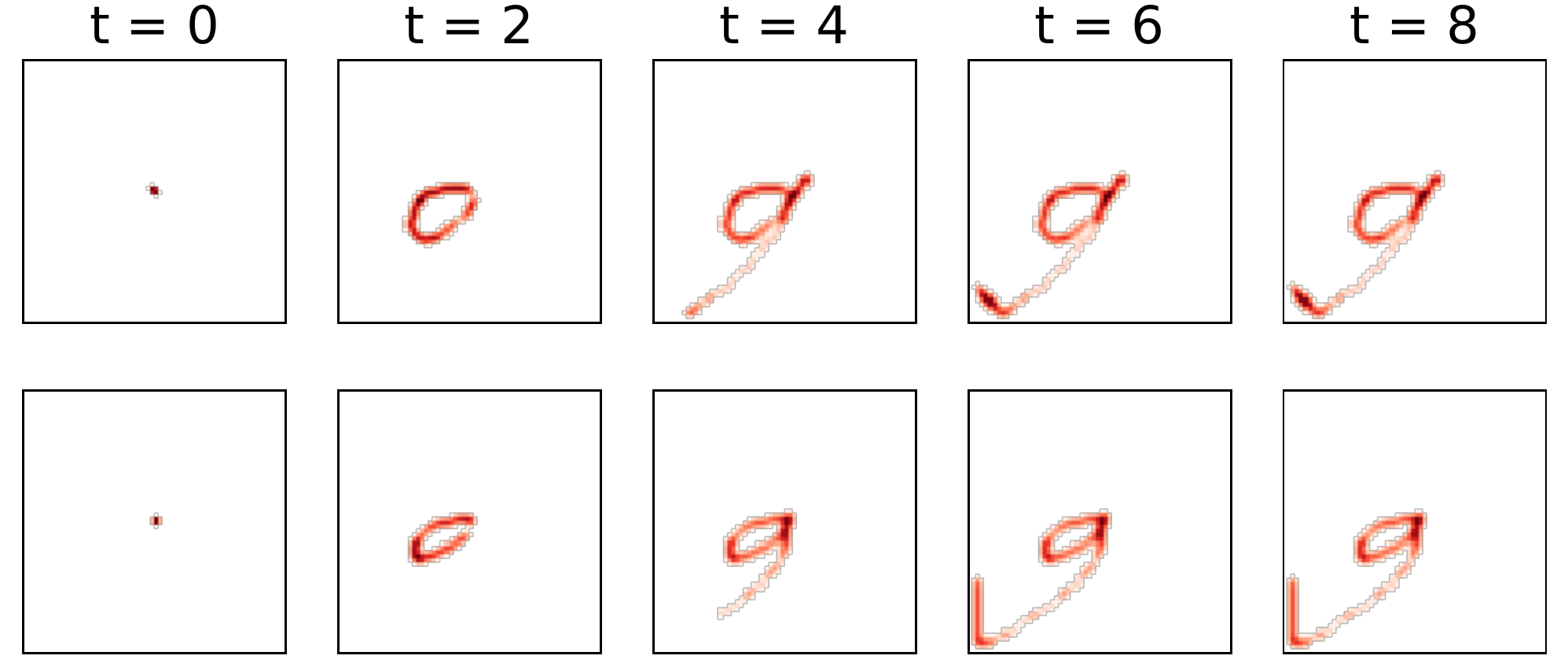}
    \caption{Examples of time series in the handwritten letters dataset corresponding to the letter "g". At each time point, $\bx_i$ corresponds to an image of the current state of the drawing. \label{f:letter-g}}
\end{figure}

To evaluate the discriminatory power of STA with real data, we use a publicly available dataset of handwritten letters where the position of a pen are tracked in time \citep{williams}. We subsample the data both spatially and temporally so as to keep 10 time points of (64$\times$64) images for each time series. Each image can thus be seen as a screenshot at a certain time during the writing motion. Figure \ref{f:letter-g} shows an example of two data samples of the letter "g".
\begin{figure}[t]
    \centering
	\includegraphics[width=0.7\linewidth]{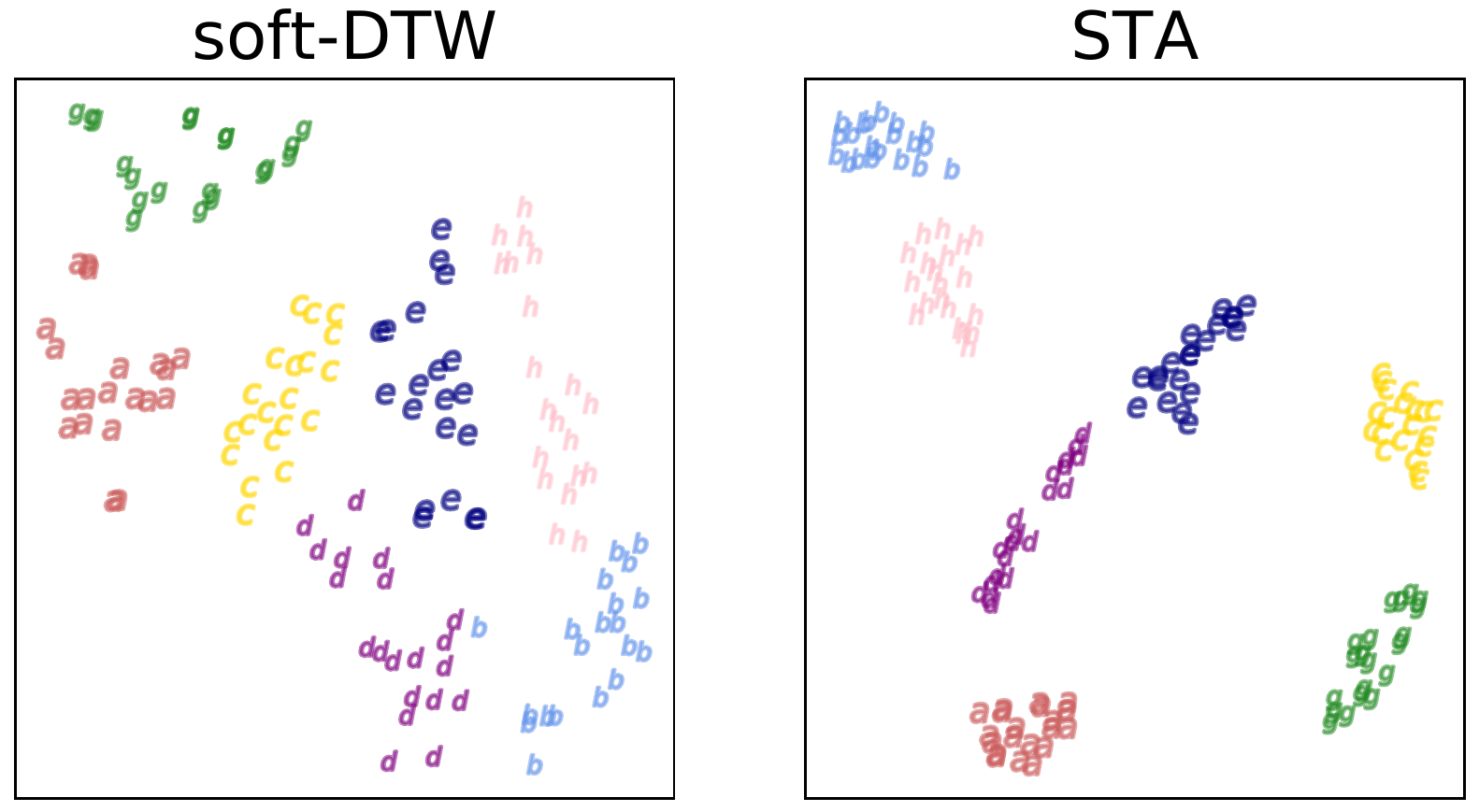}
	\caption{tSNE embeddings of the data.
		STA (proposed) captures spatial variability. \label{f:tsne}}
\end{figure}
We consider clustering 140 samples of 7 different letters -- `a` to `h` -- (20 samples per letter; 'f' was not collected in the data) with a t-SNE embedding using STA as a dissimilarity function. To speed up computation, we compute all pairwise $\delta$ distances between all images of all samples on multi-platform GPUs. Carrying out STA afterwards amounts to finding optimal assignments independently for each pair of time series. We compare STA with soft-DTW with the same $\beta = 0.1$. Figure \ref{f:tsne} shows that the choice of the dissimilarity is crucial: the spatial variability captured by the Wasserstein divergence is key to accurately discriminate between the samples. In this experiment, we noticed that the choice of $\beta$ almost did not affect the results. Given that all letters were written by the same person, all motions have similar speeds. Results with various values of $\beta$ are displayed in the supplementary materials.
%

% !TEX root = ../main.tex

\section{Conclusion}
Spatio-temporal data can differ in amplitude and in spatio-temporal structure. Our contributions are twofold. First, we showed that regularized Dynamic time warping is sensitive to temporal variability. Second, we proposed to combine an unbalanced Optimal transport divergence with soft dynamic time warping to define a dissimilarity for spatio-temporal data. The performance of our experiments on simulations and real data confirm our findings and show that our method can identify meaningful spatio-temporal clusters.

\subsection*{Acknowledgements}
MC and HJ acknowledge the support of a chaire
d’excellence de l’IDEX Paris Saclay. AG and HJ was supported by the European Research Council Starting
Grant SLAB ERC-YStG-676943.

\bibliographystyle{apalike}
\bibliography{references}

\begin{thebibliography}{}

\bibitem[Bellman, 1952]{bellman}
Bellman, R. (1952).
\newblock On the theory of dynamic programming.
\newblock In {\em Proceedings of the National Academy of Sciences}, volume~38,
  page 716–719.

\bibitem[Chizat et~al., 2017]{chizat17}
Chizat, L., Peyr{\'e}, G., Schmitzer, B., and Vialard, F.-X. (2017).
\newblock {S}caling {A}lgorithms for {U}nbalanced {T}ransport {P}roblems.
\newblock {\em arXiv:1607.05816 [math.OC]}.

\bibitem[Cuturi, 2011]{cuturi11}
Cuturi, M. (2011).
\newblock Fast global alignment kernels.
\newblock In {\em {Proceedings of the 28th International Conference on
  International Conference on Machine Learning}}, {ICML'11}, pages 929--936,
  USA. Omnipress.

\bibitem[Cuturi, 2013]{cuturi13}
Cuturi, M. (2013).
\newblock {Sinkhorn Distances: Lightspeed Computation of Optimal Transport}.
\newblock In {\em Neural Information Processing Systems}.

\bibitem[Cuturi and Blondel, 2017]{cuturi17}
Cuturi, M. and Blondel, M. (2017).
\newblock Soft-dtw: a differentiable loss function for time-series.
\newblock In {\em International Conference on Machine Learning}.

\bibitem[{Doucet} et~al., 2002]{doucet02}
{Doucet}, A., {Vo}, B.~., {Andrieu}, C., and {Davy}, M. (2002).
\newblock Particle filtering for multi-target tracking and sensor management.
\newblock In {\em Proceedings of the Fifth International Conference on
  Information Fusion.}, volume~1, pages 474--481 vol.1.

\bibitem[Feydy et~al., 2017]{feydy17}
Feydy, J., Charlier, B., Vialard, F.-X., and Peyr{\'e}, G. (2017).
\newblock Optimal transport for diffeomorphic registration.
\newblock pages 291--299.

\bibitem[Feydy et~al., 2018]{feydy19}
Feydy, J., Séjourné, T., Vialard, F.-X., Amari, S.-i., Trouvé, A., and
  Peyré, G. (2018).
\newblock Interpolating between optimal transport and mmd using sinkhorn
  divergences.
\newblock In {\em Proceedings of the Twenty-Second International Conference on
  Artificial Intelligence and Statistics}.

\bibitem[Genevay et~al., 2018]{genevay18}
Genevay, A., Peyre, G., and Cuturi, M. (2018).
\newblock Learning generative models with sinkhorn divergences.
\newblock In Storkey, A. and Perez-Cruz, F., editors, {\em Proceedings of the
  Twenty-First International Conference on Artificial Intelligence and
  Statistics}, volume~84 of {\em Proceedings of Machine Learning Research},
  pages 1608--1617. PMLR.

\bibitem[Gramfort et~al., 2011]{gramfort-etal:2011}
Gramfort, A., Papadopoulo, T., Baillet, S., and Clerc, M. (2011).
\newblock Tracking cortical activity from {M/EEG} using graph cuts with
  spatiotemporal constraints.
\newblock {\em NeuroImage}, 54(3):1930 -- 1941.

\bibitem[Janati et~al., 2019]{janati19}
Janati, H., Cuturi, M., and Gramfort, A. (2019).
\newblock Wasserstein regularization for sparse multi-task regression.
\newblock In {\em Proceedings of the Twenty-First International Conference on
  Artificial Intelligence and Statistics}, volume~89 of {\em Proceedings of
  Machine Learning Research}. PMLR.

\bibitem[Kantorovic, 1942]{doklady}
Kantorovic, L. (1942).
\newblock {On the translocation of masses}.
\newblock {\em C.R. Acad. Sci. URSS}.

\bibitem[Maaten and Hinton, 2008]{maaten08}
Maaten, L. and Hinton, G. (2008).
\newblock Visualizing high-dimensional data using t-sne.
\newblock {\em Journal of Machine Learning Research}, 9:2579--2605.

\bibitem[Pedregosa et~al., 2011]{scikit-learn}
Pedregosa, F., Varoquaux, G., Gramfort, A., Michel, V., Thirion, B., Grisel,
  O., Blondel, M., Prettenhofer, P., Weiss, R., Dubourg, V., Vanderplas, J.,
  Passos, A., Cournapeau, D., Brucher, M., Perrot, M., and Duchesnay, E.
  (2011).
\newblock Scikit-learn: Machine learning in {P}ython.
\newblock {\em Journal of Machine Learning Research}, 12:2825--2830.

\bibitem[{Peyr{\'e}} and {Cuturi}, 2018]{otbook}
{Peyr{\'e}}, G. and {Cuturi}, M. (2018).
\newblock {Computational Optimal Transport}.
\newblock {\em arXiv e-prints}.

\bibitem[Saigo et~al., 2004]{saigo04}
Saigo, H., Jean-Philippe, Vert, Ueda, N., and Akutsu, T. (2004).
\newblock Protein homology detection using string alignment kernels.
\newblock {\em Bioinformatics}, 20(11):1682–1689.

\bibitem[{Sakoe} and {Chiba}, 1978]{sakoe78}
{Sakoe}, H. and {Chiba}, S. (1978).
\newblock Dynamic programming algorithm optimization for spoken word
  recognition.
\newblock {\em IEEE Transactions on Acoustics, Speech, and Signal Processing},
  26(1):43--49.

\bibitem[Sokol, 1983]{sokoi}
Sokol, S. (1983).
\newblock Abnormal evoked potential latencies in amblyopia.
\newblock {\em The British journal of ophthalmology}, 67(5):310--314.

\bibitem[Solomon et~al., 2015]{solomon15}
Solomon, J., de~Goes, F., Peyr{\'e}, G., Cuturi, M., Butscher, A., Nguyen, A.,
  Du, T., and Guibas, L. (2015).
\newblock Convolutional {Wasserstein} distances: Efficient optimal
  transportation on geometric domains.
\newblock {\em ACM Trans. Graph.}, 34(4):66:1--66:11.

\bibitem[Stanley, 2011]{stanley11}
Stanley, R.~P. (2011).
\newblock {\em Enumerative Combinatorics: Volume 1}.
\newblock Cambridge University Press, New York, NY, USA, 2nd edition.

\bibitem[Taxi and Commission, 2019]{taxi}
Taxi, N. Y. N. Y.~. and Commission, L. (2019).
\newblock New york city taxi trip data, 2009-2018.

\bibitem[Thorpe et~al., 2017]{thorpe17}
Thorpe, M., Park, S., Kolouri, S., Rohde, G.~K., and Slep{\v c}ev, D. (2017).
\newblock A transportation l(p) distance for signal analysis.
\newblock {\em Journal of mathematical imaging and vision}, 59(2):187--210.

\bibitem[Whelan et~al., 2010]{whelan10}
Whelan, R., Lonergan, R., Kiiski, H., Nolan, H., Kinsella, K., Bramham, J.,
  O'Brien, M., Reilly, R., Hutchinson, M., and Tubridy, N. (2010).
\newblock A high-density erp study reveals latency, amplitude, and
  topographical differences in multiple sclerosis patients versus controls.
\newblock {\em Clinical neurophysiology : official journal of the International
  Federation of Clinical Neurophysiology}, 121:1420--6.

\bibitem[Williams et~al., 2006]{williams}
Williams, B., M.Toussaint, and Storkey., A. (2006).
\newblock Extracting motion primitives from natural handwriting data.
\newblock In {\em ICANN}, volume~2, page 634–643.

\end{thebibliography}

%--------------- Supplementary 
\iftoggle{supplementary}{
	\cleardoublepage
	% !TEX root = ../main.tex

\appendix
\onecolumn
\section{Proofs}
\subsection{Proof of proposition \ref{prop:kshift}}
Let $\bx \in \bbR^{T}$ be a univariate time series. Using the definitions of section \ref{s:softdtw}, proposition \ref{prop:kshift} reads:
\begin{sprop}
	\label{sprop:kshift}
	 Let $k \in \intset{1, T - 1}$, let $m = \on(\bx)$ and $m' = T - \off(\bx)$. Let $\mu = \min_{i,j}\{\Delta(\bx,~\bx)_{ij}~| \Delta(\bx,~\bx)_{ij}~>~0\}$. If   $0 < \beta \leq \frac{\mu}{\log(3TD_{T, T})}$ :
\begin{equation}
    \label{seq:kshift}
\sdtw(\bx, \bx_{+_k}) -  \sdtw(\bx, \bx)  \geq
\beta \log \left( \frac{ D_{m, m}  D_{m' , m'} }{ D_{m + k, m}  D_{m' - k , m'} } \right) - \frac{\beta}{3T}
\end{equation}
\end{sprop}
\proof
Let us remind that given a pairwise distance matrix $\Delta(\bx, \by)$, the soft-DTW dissimilarity is defined as
$\sdtw(\bx, \by) = - \beta \log\left(\sum_{A \in \cA_{T, T}} e^{-\frac{\langle A, \Delta(\bx, \by)\rangle}{\beta}}\right)$.
The set of all possible costs can be written: $C = \{ \langle A, \Delta(\bx, \by)\rangle, A \in \cA_{T, T}\}$. Dropping duplicates, let $d_0 < d_1, \dots, < d_{G}$ denote all unique values in $C$. And finally let $n_i$ be the number of alignments $A$ such that $\langle A, \Delta(\bx, \by)\rangle = d_i$. We have:
\begin{equation}
    \label{seq:sdtw-rewritten}
\sdtw(\bx, \by) = - \beta \log\left(\sum_{A  \in \cA_{T, T}} e^{-\frac{\langle A, \Delta(\bx, \by)\rangle}{\beta}}\right) = - \beta \log\left(\sum_{i=0}^G n_i e^{-\frac{d_i}{\beta}}\right) \enspace .
\end{equation}

When $\by = \bx$, we have $d_0 = 0$. Isolating the first element of the sum we get:

\begin{equation}
    \label{seq:sdtw-kshift0}
\sdtw(\bx, {\bx}) =  - \beta \log(n_0)  - \beta\log\left(1 + \sum_{i=1}^G \frac{n_i}{n_0} e^{-\frac{d_i}{\beta}}\right) \leq - \beta\log(n_0) \enspace .
\end{equation}

Similarly, when $\by$ is temporally k-shifted with respect to $\bx$, we also have $d_0 = 0$. Adding an exponent $'$ on terms that depend on the time series ${\bx_+}_k$, we have:

\begin{align}
\sdtw(\bx, {\bx_+}_k) =  - \beta \log(n_0')  -  \beta\log\left(1 + \sum_{i=1}^G \frac{n_i'}{n_0'} e^{-\frac{d_i'}{\beta}}\right) \geq - \beta\log(n_0') - \beta \sum_{i=1}^G \frac{n_i'}{n_0'} e^{-\frac{d_i'}{\beta}}
\nonumber \\
\geq - \beta\log(n_0') - \beta D_{T, T} e^{-\frac{d_1'}{\beta}}     \label{seq:sdtw-kshift1}
\end{align}

However, since the set of values taken by $\Delta(\bx, \bx)$ and $\Delta(\bx, {\bx_+}_k)$ are the same, we have $d_i=d_i'$ (but $n_i \neq n_i'$ apriori) and the assumption on $\beta$ provides:

\begin{align}
 & \beta  \leq \frac{\mu}{\log(3TD_{T,T})} \nonumber \\
        & \Rightarrow \beta \leq \frac{d_1}{\log(3TD_{T,T})} \nonumber \\
        & \Rightarrow e^{\frac{-d_1'}{\beta}} \leq \frac{1}{3TD_{T, T}} \nonumber \\
        & \Rightarrow -\beta D_{T, T} e^{\frac{-d_1'}{\beta}} \geq - \frac{\beta}{3T} \label{seq:betabound}
\end{align}

Combining \eqref{seq:sdtw-kshift0}, \eqref{seq:sdtw-kshift1} and \eqref{seq:betabound} leads to:
\begin{equation}
    \label{seq:kshift-n0}
    \sdtw(\bx, {\bx_+}_k) - \sdtw(\bx, \bx) \geq \beta \log\left(\frac{n_0}{n_0'}\right) - \frac{\beta}{3T}
\end{equation}

Now let's develop the term $\frac{n_0}{n_0'}$. 
$n_0'$ corresponds to the number of equivalent alignments with 0 cost which can be given by $D_{\on(\bx) , \on(\by) } \Omega  D_{T - \off(\bx) , T - \off(\by)}$, where $\Omega$ is the number of 0 cost alignments within the cross product of the fluctuation sets. However, temporal shifts do not change $\Omega$ but only change the outermost sets. For instance, considering the example of Figure~\ref{f:shiftexample} one can see that only rectangles outside the fluctuation set are affected. Therefore, $\Omega$ cancels out in $\frac{n_0}{n_0'}$ and we get the desired bound.
\qed

\subsection{Proof of proposition \ref{prop:delannoy-ineq}}

For the sake of completeness, we start this section by reminding some key results on Delannoy numbers.
\subsection*{Delannoy numbers}

We re-define the Delannoy sequence starting from $m=n=1$ so as to correspond to the number of chronological alignments  in the (1, 1) $\to (m, n) $ lattice: $\card(\cA_{m, n}) = D_{m, n}$.
\begin{sdefinition}[Delannoy sequence]
	\label{sdef:delannoy}
	The Delannoy number $D_{m, n}$ corresponds to the number of paths from $(1, 1)$ to $(m, n)$ in a $(m \times n)$ lattice where only  $\rightarrow, \downarrow, \searrow$ movements are allowed. It can also be defined with the recursion $\forall m, n \in \bbN^{\star}$:
	\begin{align}
	& D_{1, n} = D_{m, 1} = 1 \\
	&  D_{m + 1, n + 1} = D_{m , n + 1} + D_{m + 1, n} + D_{m , n }  \label{seq:delannoy-rec} \enspace .
	\end{align}
\end{sdefinition}

The central (or diagonal) Delannoy numbers $D_m = D_{m, m}$ verifiy an intersting 2-stages recursion equation:

\begin{sprop}[\cite{stanley11}]
	\label{sprop:central-recursion}
	For $m \geq 2$:
	\begin{equation}
	\label{seq:central-recursion}
	m D_{m + 1} = (6m - 3) D_{m} - (m - 1)D_{m-1}
	\end{equation}
\end{sprop}

\proofsketch The proof of \citet{stanley11} is based on the closed form expression of Delannoy numbers: $D_m = \sum_{k=1}^m \binom{m, k}{m + k, k}$ and the generating function $\sum_{m=1}^\infty D_mx^n = \frac{1}{\sqrt{1- 6x + x^2}}$. Taking the derivative of the power series yields the desired recursion equation.

\begin{slemma}[Bounded growth - Lemma \ref{lem:growth}]
	\label{slem:growth}
	Let $c = 1 + \sqrt{2}$ and $m \geq 2$. The sequence of central Delannoy numbers $D_m = D_{m, m}$ verifies:
	\begin{equation}
	\label{seq:growth}
P(m):	D_{m+1} \leq c^2 D_{m}
	\end{equation}
\end{slemma}

\proof Proof by induction. For $m = 1$, we have $D_2=3 \leq (3 + 2\sqrt{2}) = c^2 = c^2 D_1$. Let $m \geq 2$ and assume $P(m)$ is true.
From \eqref{seq:central-recursion} and $P(m)$ we have:
\begin{align}
(m + 1)D_{m+2} &= (6m + 3)D_{m + 1} - mD_{m} \\
								& \leq (6m + 3 - \frac{m}{c^2}) D_{m + 1} \\
								& \leq (6 - \frac{1}{c^2}) m D_{m + 1} \\
								& \leq (6 - \frac{1}{c^2}) (m + 1) D_{m + 1}
\end{align} 
And we also have $1 / c^2 = \frac{1}{3 + 2\sqrt{2}} = 3 - 2\sqrt{2}$, hence $6 - \frac{1}{c^2} = c^2$; we have $P(m+1)$. \qed

\subsection*{Proof of proposition \ref{prop:delannoy-ineq}}
Proposition \ref{prop:delannoy-ineq} is our most technical contribution, its demonstration requires considerable care. Similarly to bounded growth Lemma \ref{slem:growth}, we would like to bound the off-diagonal Delannoy numbers with their closest diagonal numbers with a bound depending on $k$. We do so incrementally by comparing the off-diagonal number $D_{m, m + k}$ with $D_{m, m + k -1}$ and $D_{m + 1, m + k}$. The proposition states:

\begin{sprop}[Proposition \ref{prop:delannoy-ineq}]
	\label{sprop:delannoy-ineq}
	Let $c = 1 + \sqrt{2}$.  $\forall m, k \in \bbN^\star$:
	\begin{align}
	\label{seq:delannoy-ineq}
	A(m, k): \quad &D_{m, m + k}  \leq c \Phi_{m, k} D_{m, m + k - 1} \\ 
	B(m, k): \quad &c \Psi_{m, k} D_{m, m + k}  \leq D_{m + 1, m + k}
	\end{align}
	Where
	\[
	\left\{	 \begin{array}{l} 
	\Phi_{m, k} = 1 - \frac{ (1 - \frac{1}{c}) (k - 1) + \frac{1}{c}}{m + k - 1} \\
	\Psi_{m, k} = 1 + \frac{(1 - \frac{1}{c}) (k-1)}{m} 
	\end{array}
	\right.
	\]
\end{sprop}
It is noteworthy that -- since $1 - 1/c = 2 - \sqrt{2} > 0 $ -- we have for all $m, k$ $\Phi_{m, k}  \leq 1$ and $\Psi_{m, k} \geq 1$. When both $\Psi$ and $\Phi$ are constant and equal to 1, we get two constant bounds equal to $c$. The role of $\Phi$ and $\Psi$ is to have tighter bounds when $k$ increases.
The demonstration is based on an induction reasoning on $m$. That is, we would like to show for all $m$ the statement: $P(m):  (\forall k \geq 1)\ A(m, k)$ and $B(m, k)$.
To assist the reader, we visualize the proof on Figure~\ref{f:proof} which describes all the steps of the induction.  For the sake of clarity, we isolate the following technical Lemma before proving the proposition.

\begin{slemma}
	\label{slem:technical}
	Let $c = 1 + \sqrt{2}$ and $m, k \geq 1$. The sequences $\Phi$ and $\Psi$ verify the inequalities:
	\begin{equation}
	\label{seq:technical}
		c\Psi_{m, k + 1} \Phi_{m, k + 1}  \leq \left(\frac{1}{c} +  \Psi_{m , k}  + \Phi_{m, k + 1}  \right) \leq c \Phi_{m +1, k} \Psi_{m, k}
	\end{equation}
\end{slemma}

\proof
First, a notation to make calculations easier, let $\alpha = 1 - \frac{1}{c}$. Then we have:
\[
\left\{	 \begin{array}{l} 
\Phi_{m, k} = 1 - \frac{ a (k - 1) + \frac{1}{c}}{m + k - 1} \\
\Psi_{m, k} = 1 + \frac{a(k - 1)}{m} 
\end{array}
\right.
\]
The middle term can be written using $2 + \frac{1}{c} = c$,
\begin{align*}
\frac{1}{c} +  \Psi_{m , k}  + \Phi_{m, k + 1} &= 2 + \frac{1}{c}+ \frac{a(k-1)}{m}   - \frac{ a k + \frac{1}{c}}{m + k}  \\
& = c  + \frac{a(k - 1)}{m} - \frac{ a k + \frac{1}{c}}{m + k}  \enspace.
\end{align*}

Let's start by proving the right inequality. 
\begin{description}
	\item[1. Right inequality:]
The right side can be written:
\begin{equation}
 c\Phi_{m +1, k} \Psi_{m, k} = c + c\left[\frac{ak}{m} - \frac{ a (k - 1) + \frac{1}{c}}{m + k}   - \frac{a(k-1)}{m}  \frac{ \left(a (k - 1) + \frac{1}{c}\right)}{m + k} \right] 
 \end{equation}

The inequality we want to prove is equivalent to, dropping the first $c$:
For all $m, k \geq 1$:
\begin{align*}
 & \frac{a(k - 1)}{m} - \frac{ a k + \frac{1}{c}}{m + k}  \leq  c\left[\frac{a(k-1)}{m}  - \frac{ a (k - 1) + \frac{1}{c}}{m + k}  - \frac{a(k-1)}{m}  \frac{ \left(a (k - 1) + \frac{1}{c}\right)}{m + k} \right] \\
 & \Leftrightarrow a(k-1)(m + k) - m(ak + \frac{1}{c}) \leq c\left[ a(k-1)(m + k) - m\left( a(k-1) + \frac{1}{c}\right) - a(k-1)\left(a(k-1) + \frac{1}{c}\right) \right] \\
 &\Leftrightarrow akm + a k^2 - am - ak - mak - \frac{m}{c} \leq c \left[ akm + ak^2 - ma - ak - akm + ma - \frac{m}{c} - a^2(k-1)^2 - \frac{a}{c} k + \frac{a}{c} \right] \\
 &\Leftrightarrow a\left(c - a c - 1\right) k^2 + ac \left(2 a - 1\right) k + m\left(a + \frac{1}{c} - 1\right) + a - a^2 c \geq 0
\end{align*}

However, $c - ac - 1 = 0$ and $ a + \frac{1}{c} - 1 = 0$. Thus, the left side above gives rise to an affine function $f$ in $k$ defined as: 
$f(k) = ac( 2 a - 1)k + a - a^2 c $ that verifies  $f(1) = ac ( 2a - 1) + a - a^2 c = 0$, and since its slope $ac( 2 a - 1) = a (2c - 3) = a(\sqrt{2} - 1) > 0$, we have $f(k) \geq 0, \quad \forall k \geq 1$. Therefore, since all inductions above are equivalent to each other, the right inequality is proven.
 
 \item[2. Left inequality:]
 
 Similarly,  the left side can be written:
\begin{equation}
 c\Phi_{m, k + 1} \Psi_{m, k + 1} = c +c  \left[ \frac{ak}{m}   - \frac{ a k + \frac{1}{c}}{m + k}  - \frac{ak}{m}  \frac{ \left(a k + \frac{1}{c}\right)}{m + k} \right]
\end{equation}

Again c cancels out, and the inequality is equivalent to, for all $m, k \geq 1$:
\begin{align*}
&  \frac{a(k - 1)}{m} - \frac{ a k + \frac{1}{c}}{m + k}  \geq c  \left[ \frac{ak}{m}   - \frac{ a k + \frac{1}{c}}{m + k}  - \frac{ak}{m}  \frac{ \left(a k + \frac{1}{c}\right)}{m + k} \right]\\
& \Leftrightarrow  akm + a k^2 - am - ak - mak - \frac{m}{c} \geq c  \left[ akm + ak^2 - akm - \frac{m}{c} - a^2 k^2 - \frac{ak}{c}\right] \\
& \Leftrightarrow a\left(c - a c - 1\right) k^2  + m\left(a + \frac{1}{c} - 1\right) \geq 0
\end{align*}

However, $c - ac - 1 = 0$ and $ a + \frac{1}{c} - 1 = 0$. Thus, we indeed have the last inequality. Therefore, since all inductions above are equivalent to each other, the left inequality is proven.
\qed

\end{description}

\paragraph{Proof of proposition \ref{prop:delannoy-ineq}}
We can now describe our induction proof. We would like to show for all $m$ the statement: $P(m):  (\forall k \geq 1)\ A(m, k)$ and $B(m, k)$.
\begin{description}
\item[0. intialization step] For $m=1$, on one hand we have for all $k\geq 1: D_{1, k} = 1$  and $c\Phi_{1, k} = 1 + \frac{c - 2}{k} = 1 + \frac{\sqrt{2} - 1}{k} \geq 1$, thus we have $A(1, k)\ \forall k$.
On the other hand, one can easily show that $D_{2, 1 + k} = 2 k + 1$ and that $c\Psi_{1, k} = (c - 1) k + 1 = \sqrt{2} k + 1 \leq 2k + 1$, since $D_{1, k + 1 } = 1$, we have $B(1, k)\ \forall k$.

\item[1. induction step (on m)]. Let $m \geq 2$ and assume $A(m, k)$ and $B(m, k)$ are true for all $k\geq1$. We first start by proving $A(m + 1, k)$ for any  $k\geq 1$.
\begin{description}
	\item[1.1 $A(m, k)$ and $B(m, k) \, (\forall k) \Rightarrow A(m + 1, k) \, (\forall k)$:] 
	We show this directly for any $k \geq 1$.
	Using the recursive definition of Delannoy numbers \eqref{seq:central-recursion} applied to left side of $A(m + 1, k)$ we have:
	\begin{equation}
	\label{seq:recursion-ineq1}
	D_{m +1, m + k + 1} = D_{m + 1, m + k} + D_{m, m + k + 1} + D_{m, m + k}
	\enspace .
	\end{equation}
	Applying $A(m, k + 1)$ to the second term of the right side we get: $D_{m, m + k + 1}  \leq c \Phi_{m, k + 1} D_{m, m + k}$; and applying $B(m, k)$ to the third term, we get: $ D_{m, m + k} \leq \frac{D_{m + 1, m + k}}{c \Psi_{m, k}} $.
	Which sums up to: $ D_{m +1, m + k + 1}~\leq~\left( 1 + \frac{1}{c\Psi_{m, k}} + \frac{\Phi_{m, k + 1}}{\Psi_{m , k}} \right) D_{m + 1, m + k}$. To conclude $A(m + 1, k)$, all we need is:
	\begin{equation}
	\label{seq:ineq1}
	 \left( 1 + \frac{1}{c\Psi_{m, k}} + \frac{\Phi_{m, k + 1}}{\Psi_{m , k}} \right) \leq c \Phi_{m +1, k} \enspace ,
	\end{equation}
	which follows directly from the right inequality of Lemma  \ref{slem:technical}. We have thus proven $A(m + 1, k)$ for any arbitrary $k \geq 1$.

	\item[1.2 $A(m, k)$, $A(m + 1, k)$,  $B(m, k) \, (\forall k) \Rightarrow B(m + 1, k) \, (\forall k)$:] 
	We prove the statement $B(m + 1, k) \, (\forall k)$ via an induction reasoning on $k$. 
	\begin{description}
		\item[1.2.0 initialization step (k = 1)]. For $k=1$, we have to show that $c \Psi_{m + 1, 1} D_{m + 1, m + 2} \leq D_{m+2}$. On one hand, we have $\Psi_{m + 1, 1} = 1$. On the other hand, using the recursion definition \eqref{seq:delannoy-rec} we get: $D_{m + 2} = D_{m + 1, m + 2} + D_{m +2, m  + 1} + D_{m + 1}$. And by symmetry of Delannoy numbers: $ D_{m + 2}~=~2~D_{m +1 , m +2}~+~D_{m + 1}$.
		Now using the growth lemma \ref{slem:growth} on $D_{m + 1}$, we have: $D_{m +1 , m +2} \leq \frac{c^2 - 1}{2 c^2} D_{m + 2}$. Since $c = 1 + \sqrt{2}$, we have $\frac{c^2 - 1}{2 c^2}= \frac{1}{c}$ which concludes $B(m + 1, 1)$.
		\item[1.2.1 induction step (on k)]:. Let $k \geq 1$ and assume $B(m + 1, k)$ is true, let's prove that $B(m + 1, k + 1)$ is true as well.
		$B(m + 1, k + 1)$ can be written: $c \Psi_{m + 1, k + 1} D_{m +1, m + k + 2} \leq D_{m +2, m + k + 2}$. 
		Again, using the recursion definition, we have:
			\begin{equation}
		\label{seq:recursion-ineq2}
		D_{m +2, m + k + 2} = D_{m + 1, m + k + 2} + D_{m, m + k + 1} + D_{m  + 1, m + k + 1} + D_{m + 2, m + k + 1}
		\end{equation}
		Applying the already proven $A(m + 1,~k')$ (for all k') to the second member of the right side, we have: $D_{m + 1, m + k + 1} \geq \frac{D_{m + 1, m + k + 2}}{c \Phi_{m + 1, k + 1}}$. Similarly, applying the induction (on k) assumption $B(m + 1,~k)$ to the third member, we get: $D_{m + 2, m + k + 1} \geq c \Psi_{m +1 , k} D_{m +1, m + k + 1 }$.
		 Which sums up to: $D_{m +2, m + k + 2}~\geq~\left(1 +  \frac{1}{c \Phi_{m +1 , k + 1}} + \frac{\Psi_{m + 1, k}}{\Phi_{m + 1, k + 1}} \right) D_{m + 1, m + k + 2}$. To conclude $B(m+1, k+1)$, all we need is:
		 \begin{equation}
		 \label{seq:ineq2}
		 c \Psi_{m +1, k + 1} \leq \left( 1 + \frac{1}{c\Phi_{m + 1, k + 1}} + \frac{\Psi_{m +1,  k}}{\Phi_{m + 1, k + 1}} \right) 
		 \end{equation}
		Which follows directly from the left inequality of Lemma \ref{slem:technical}, where $m$ is substituted with $m + 1$.  
		Therefore, $B(m + 1, k + 1)$ is true, ending the induction proof on $k$. 
	\end{description}
\end{description}
	Hence, $B(m + 1, k)$ holds  for any $k \geq 1$, the induction on proof on $m$ is complete. \qed
\end{description}

\begin{figure}[h]
\begin{minipage}{0.5\linewidth}
    \includegraphics[width=\linewidth]{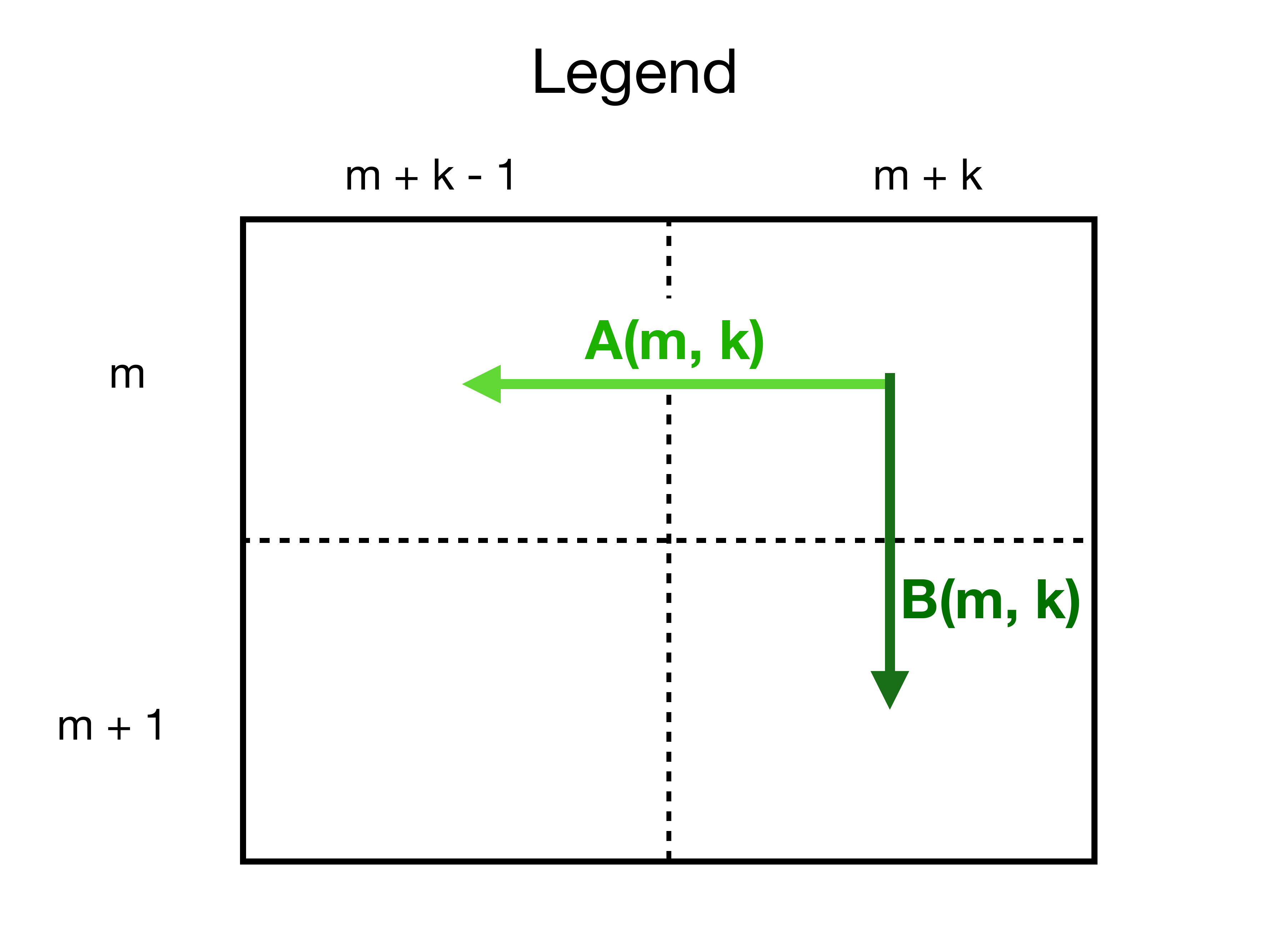}
\end{minipage}
\begin{minipage}{0.5\linewidth}
    \includegraphics[width=\linewidth]{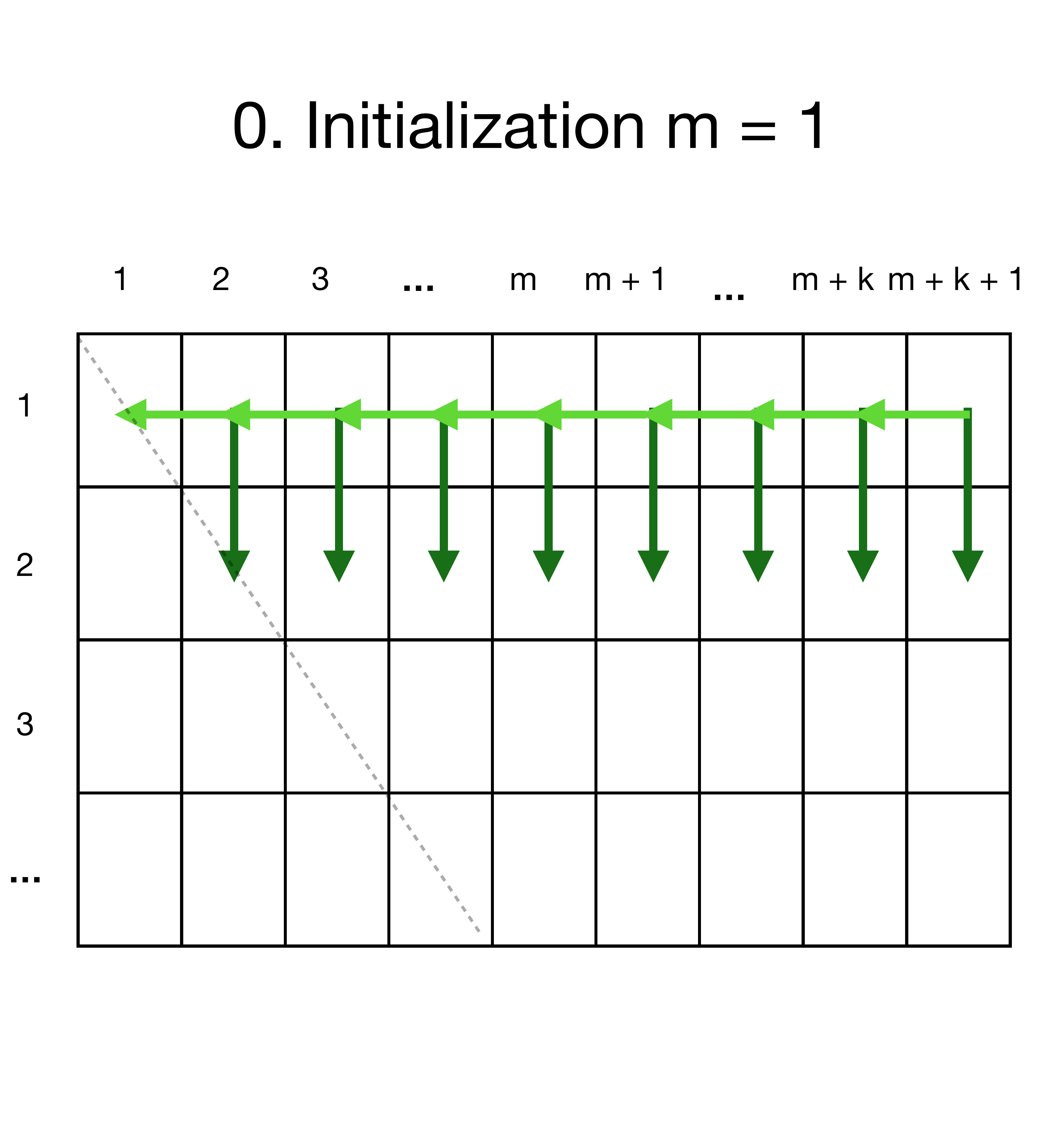}
\end{minipage}

\begin{minipage}{0.5\linewidth}
    \includegraphics[width=\linewidth]{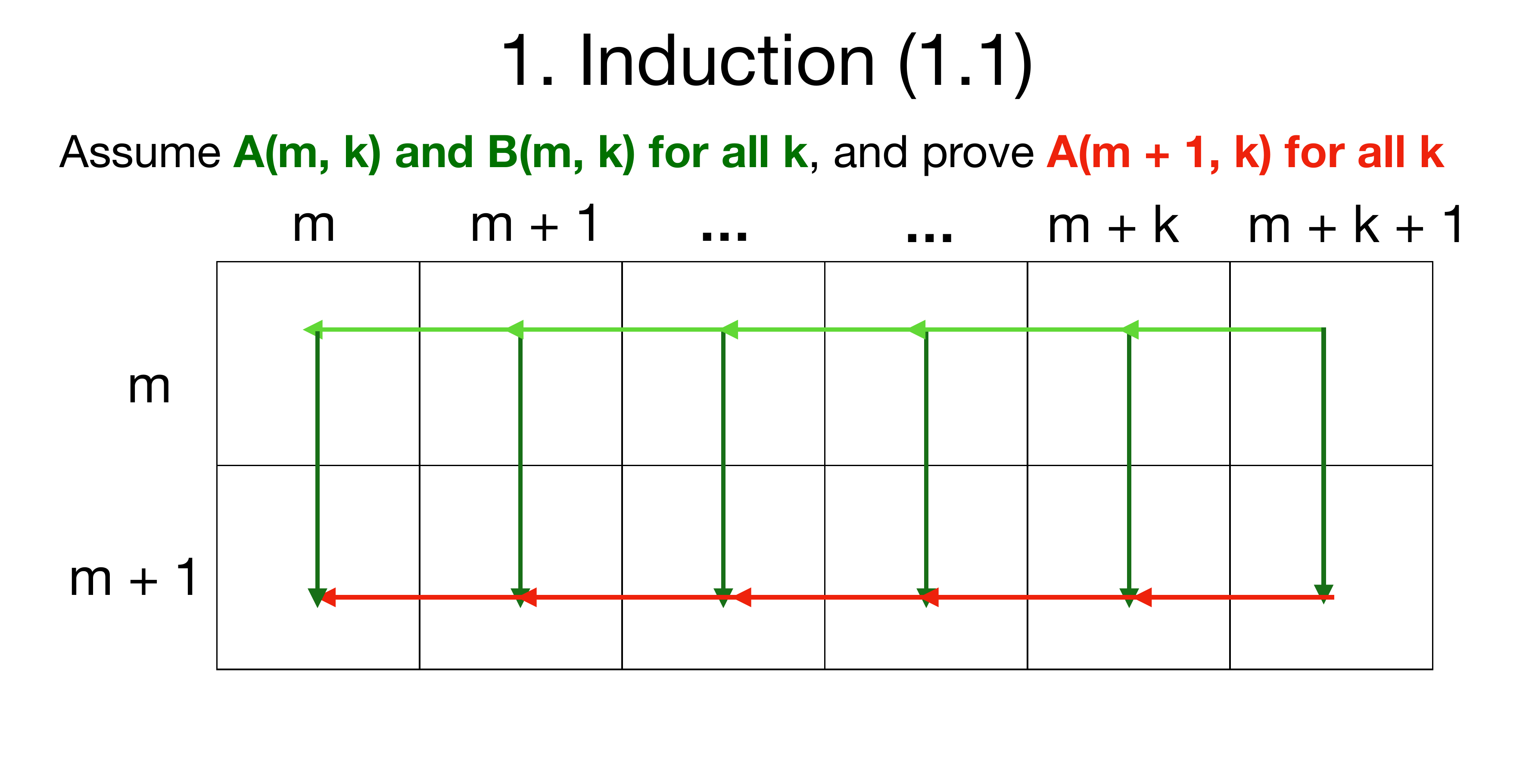}
\end{minipage}
\begin{minipage}{0.5\linewidth}
    \includegraphics[width=\linewidth]{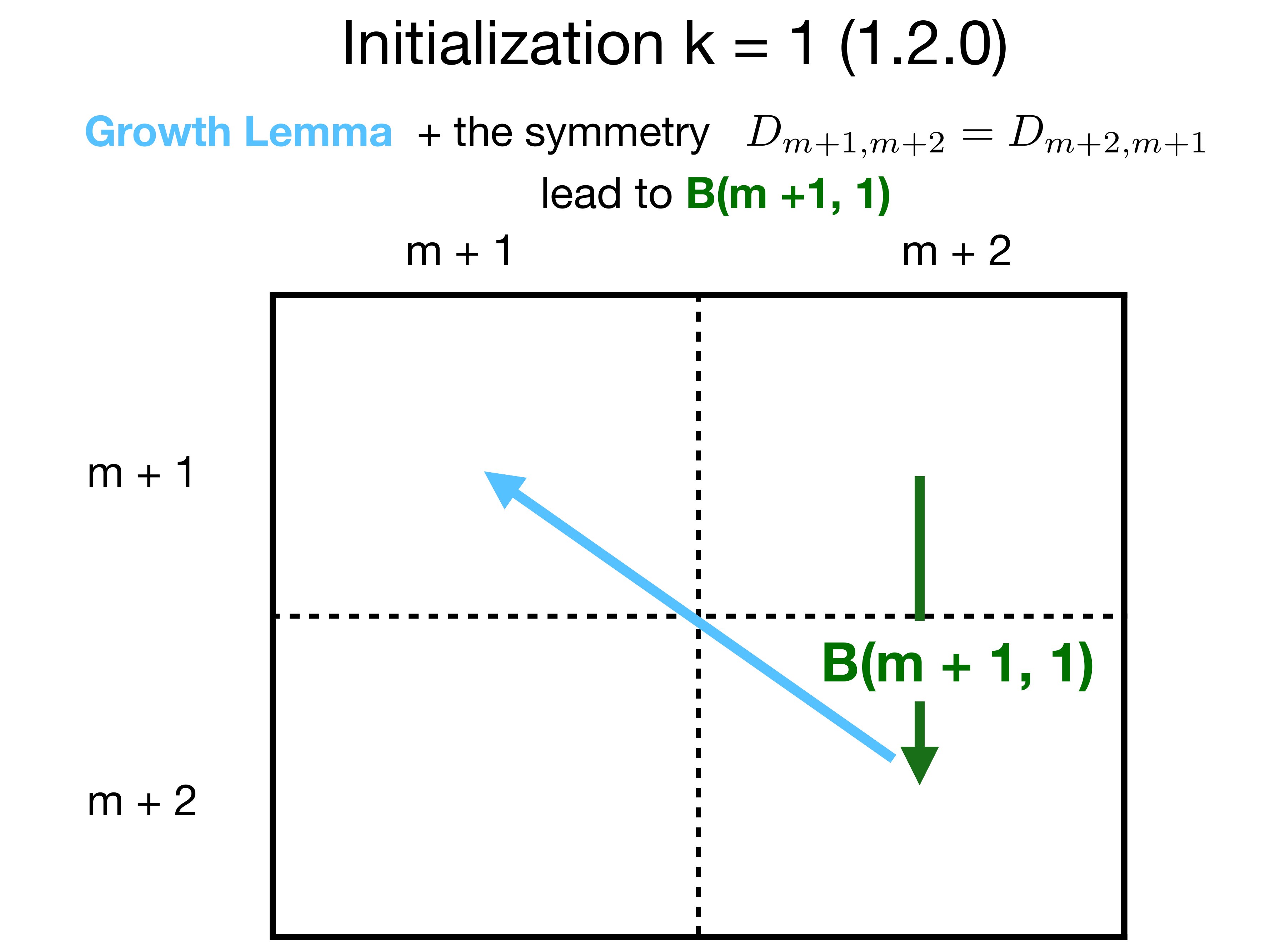}
\end{minipage}

\includegraphics[width=0.5\linewidth]{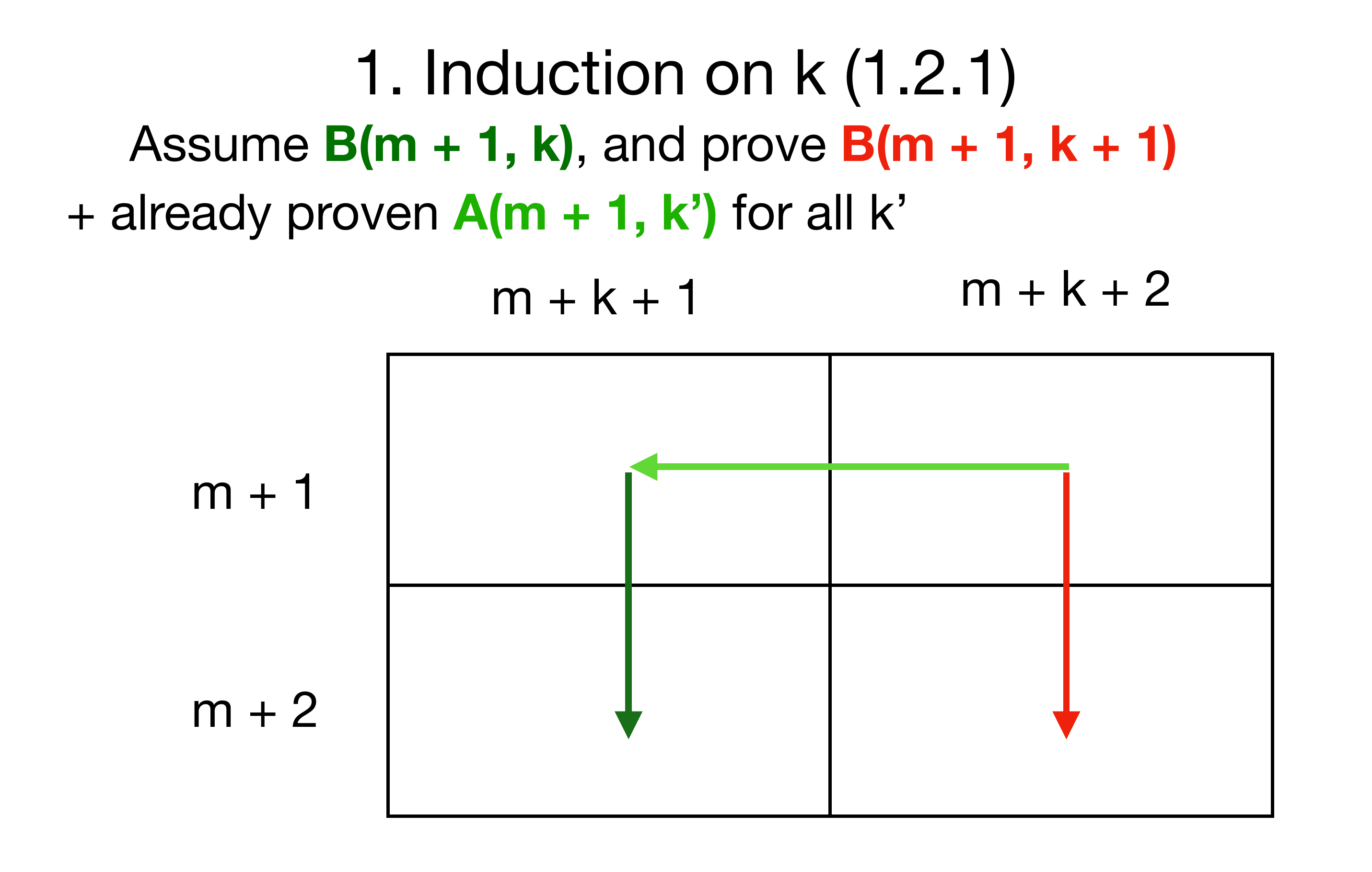}
\caption{Visualization of the proof of proposition \ref{prop:delannoy-ineq}. The key steps are 1.1 and 1.2.1, where given the top and left arrows, one must derive the right and bottom arrows. \label{f:proof}}
\end{figure}

\subsection{Other proofs}

\begin{slemma}[Bounded transported mass]
	\label{slem:planmass}
	Let $\bx, \by \in \bbR_{+}^{p}$ and $\bP_{\bx, \by} \in \bbR_+^{p\times p}$ their associated transport plan, solution of \eqref{eq:unbalanced-wasserstein}. Let $\kappa~=~\min_{i, j} e^{-\frac{M_{ij}}{\gamma}}$. We have the following bounds on the total transported mass:
	\begin{equation}
	\label{seq:planmass}
	\kappa \|\bx\|_1 \|\by\|_1 \leq \| \bP_{\bx, \by}\|^{2 + \frac{\varepsilon}{\gamma}}_1 \leq p^{2(1 + \frac{\varepsilon}{\gamma})} \|\bx\|_1 \|\by\|_1
	\end{equation}
\end{slemma}
\proof The first order optimality condition of \eqref{eq:unbalanced-wasserstein} reads for all $i, j \in \intset{1, p}$:
\begin{align}
\label{eq:KKT}
&\varepsilon \log(\bP_{ij}) - \varepsilon \log(\bK_{ij}) + \gamma \log\left(\frac{\bP_{i.}^\top\mathds 1  \bP_{.j}^\top\mathds 1}{\bx_i \by_j}\right) = 0 \\
&\Leftrightarrow \bP_{ij}^{\frac{\varepsilon}{\gamma}} \bP_{i.}^\top\mathds 1  \bP_{.j}^\top\mathds 1 = \bx_i \by_j e^{-\frac{\bM_{ij}}{\gamma}}
\end{align}
On one hand we have:
\begin{align*}
\sum_{i, j}^p \bP_{ij}^{\frac{\varepsilon}{\gamma}} \bP_{i.}^\top\mathds 1  \bP_{.j}^\top\mathds 1 
& \leq \| \bP\|^{\frac{\varepsilon}{\gamma}}_\infty \sum_{i, j}^p \bP_{i.}^\top\mathds 1  \bP_{.j}^\top\mathds 1 \\
& = \| \bP\|^{\frac{\varepsilon}{\gamma}}_\infty \|\bP\|^2_1 \\
& \leq  \|\bP\|^{2 + \frac{\varepsilon}{\gamma}}_1
\end{align*}
On the other hand, using Jensen's inequality in the second step:
\begin{align*}
\sum_{i, j}^p \bP_{ij}^{\frac{\varepsilon}{\gamma}} \bP_{i.}^\top\mathds 1  \bP_{.j}^\top\mathds 1 
& \geq  \sum_{i, j}^p  \bP_{ij}^{\frac{\varepsilon}{\gamma} + 2} \\
& \geq p^2 \left(\frac{ \sum_{i, j}  \bP_{ij} }{p^2} \right)^{\frac{\varepsilon}{\gamma} + 2} \\
& \geq  p^{-2 - 2\frac{\varepsilon}{\gamma}} \|\bP\|^{2 + \frac{\varepsilon}{\gamma}}_1
\end{align*}
 Finally, since $\kappa \leq \min_{ij} e^{-\frac{\bM_{ij}}{\gamma}} \leq 1$ and $\frac{ 2 + 2 \frac{\varepsilon}{\gamma} }{2 + \frac{\varepsilon}{\gamma}} \leq \frac{3}{2}$,we get the desired inequalities.
 \qed
\paragraph{Differentiability of $W$}
\begin{sprop}
	\label{sprop:gradientw}
	Given a fixed $\by \in \bbR^{n\times p}_+$, the unbalanced Wasserstein distance function $\bx \to W(\bx, \by)$ is smooth and its gradient is given by:
	\begin{equation}
	\label{seq:gradientw}
	\nabla_{\bx}W(\bx, \by) = \gamma (1 - {a(\bx, \by})^{-\frac{\varepsilon}{\gamma}})
	\end{equation}
	Where $\ba(\bx, \by)$ is the optimal Sinkhorn scaling, solution of the fixed point problem \eqref{eq:fixedpoint}.
\end{sprop}
\proof
As noted by \citet{feydy17}. The proof is similar to the balanced case. Indeed by applying the envelope theorem to the equivalent dual problem \eqref{eq:dualw}, one has $\nabla_{\bx}W(\bx, \by) = \nabla_{\bx}
\left(- \gamma \langle e^{-\frac{u}{\gamma}} - 1, \bx\rangle
\right)
= \gamma  \left(1 - e^{-\frac{u}{\gamma}}\right) = \gamma (1 - {a(\bx, \by)}^{- \frac{\varepsilon}{\gamma}})$ with the change of variable $\varepsilon \log(a) = u$.
\qed

\begin{sprop}
	\label{prop:stationarypoints}
	Let $\by, \bx  \in \bbR^{n\times p}_{++}$ be a stationary point of $S$ i.e $\nabla S(\bx, \by) = (\boldsymbol{0}, \boldsymbol{0})$. Then,
	$S(\bx, \by) = 0$. 
	Moreover, if $\bK$ is positive definite, then $\bx = \by$.
\end{sprop}
\proof Let $\ba, \bb, \bc, \bd$ the solutions of the fixed problems:
\begin{equation}
\label{eq:fixedpoint}
\ba = \left(\frac{\bx}{K\bb}\right)^{\omega} \quad , \quad
\bb = \left(\frac{\by}{K\ba}\right)^{\omega}, \quad
\bc = \left(\frac{\bx}{K\bc}\right)^{\omega}, \quad
\bd = \left(\frac{\by}{K\bd}\right)^{\omega}
\end{equation}
We have applying the chain rule, $\frac{1}{2}$ disappears and we get:
$\nabla_x S(\bx, \by) = \gamma (\bc^{-\frac{\varepsilon}{\gamma}} - \ba^{-\frac{\varepsilon}{\gamma}})$
and
$\nabla_y S(\bx, \by) = \gamma (\bd^{-\frac{\varepsilon}{\gamma}} - \bb^{-\frac{\varepsilon}{\gamma}})$

If $(\bx, \by)$ is a stationary point of $S$, then we immediately have $\ba = \bc$ and $\bb = \bd$. The fixed point equations lead to $\bK\ba = \bK\bb = \bK\bc = \bK\bd$. 

The transported mass between $\bx$ and $\by$ is given by:  $\|P_{\bx, \by}\|_1  = \langle \ba,  \bK\bb \rangle = \langle \bb,  \bK\ba \rangle$. Therefore, using Proposition~\ref{prop:Smass}, $S$ can be written:
\begin{align*}
S(\bx, \by) &= \frac{\varepsilon + 2\gamma}{2} (\langle  \bc,  \bK\bc \rangle + \langle \bd,  \bK\bd \rangle - 2\langle \ba,  \bK\bb \rangle) \\
&= \frac{\varepsilon + 2\gamma}{2} (\langle  \bc,  \bK\bc \rangle + \langle \bd,  \bK\bd \rangle - 2\langle \ba,  \bK\bb \rangle - \langle \bb,  \bK\ba \rangle) \\
	&=  \frac{\varepsilon + 2\gamma}{2} (\langle \bc + \bd - \ba -\bb, \bK\ba\rangle) \\
	&= 0
\end{align*}

Moreover, if $\bK$ is positive definite, $\bK\ba = \bK\bb$ leads to $\ba = \bb$ and thus $\bx = \by$.
\qed
\section{Experiments}
In this section, we provide details on the experimental settings as well as complementary results.

\subsection{Temporal shift bound}
We proved in the theoretical section that our quadratic bound holds for $0 < \beta \leq  \frac{\mu}{\log(3TD_{T,T})} $. We argue that this upper bound is too restrictive in practice. We show the empirical comparisons for different values of beta. When $\beta$ is too large, (here $\beta > 10$) one can see that the quadratic bound does not hold for large temporal shifts.
To get a tighter general bound, one must carry out a finer analysis of the remaining logsumexp terms after isolating the first large term $\frac{n_0}{n_0'}$. 

\begin{figure}
    \centering
    \begin{minipage}{0.49\linewidth}
            \includegraphics[width=\linewidth]{img/bound.pdf}
    \end{minipage}
    \begin{minipage}{0.49\linewidth}
             \includegraphics[width=\linewidth]{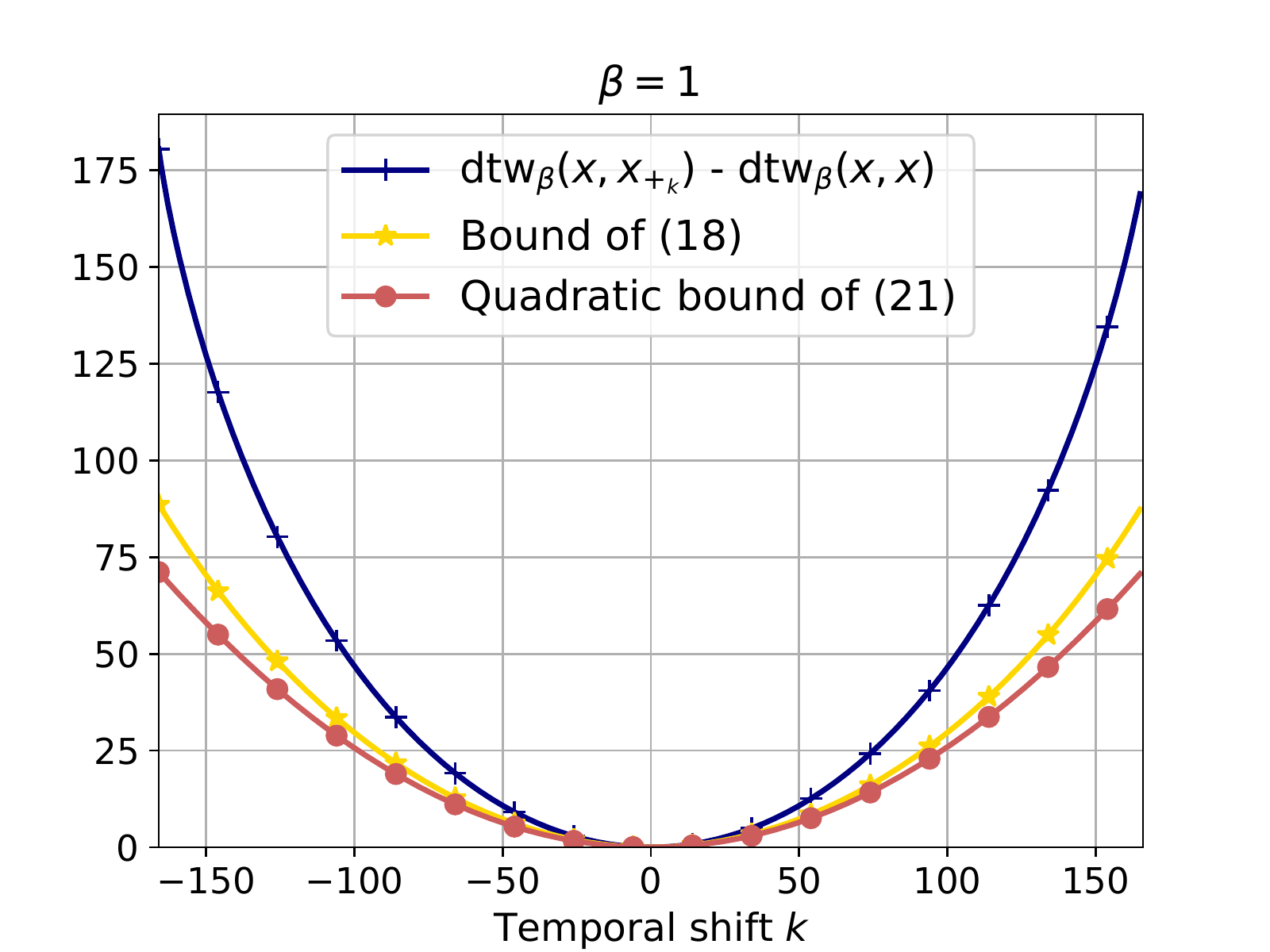}
    \end{minipage}

    \begin{minipage}{0.49\linewidth}
            \includegraphics[width=\linewidth]{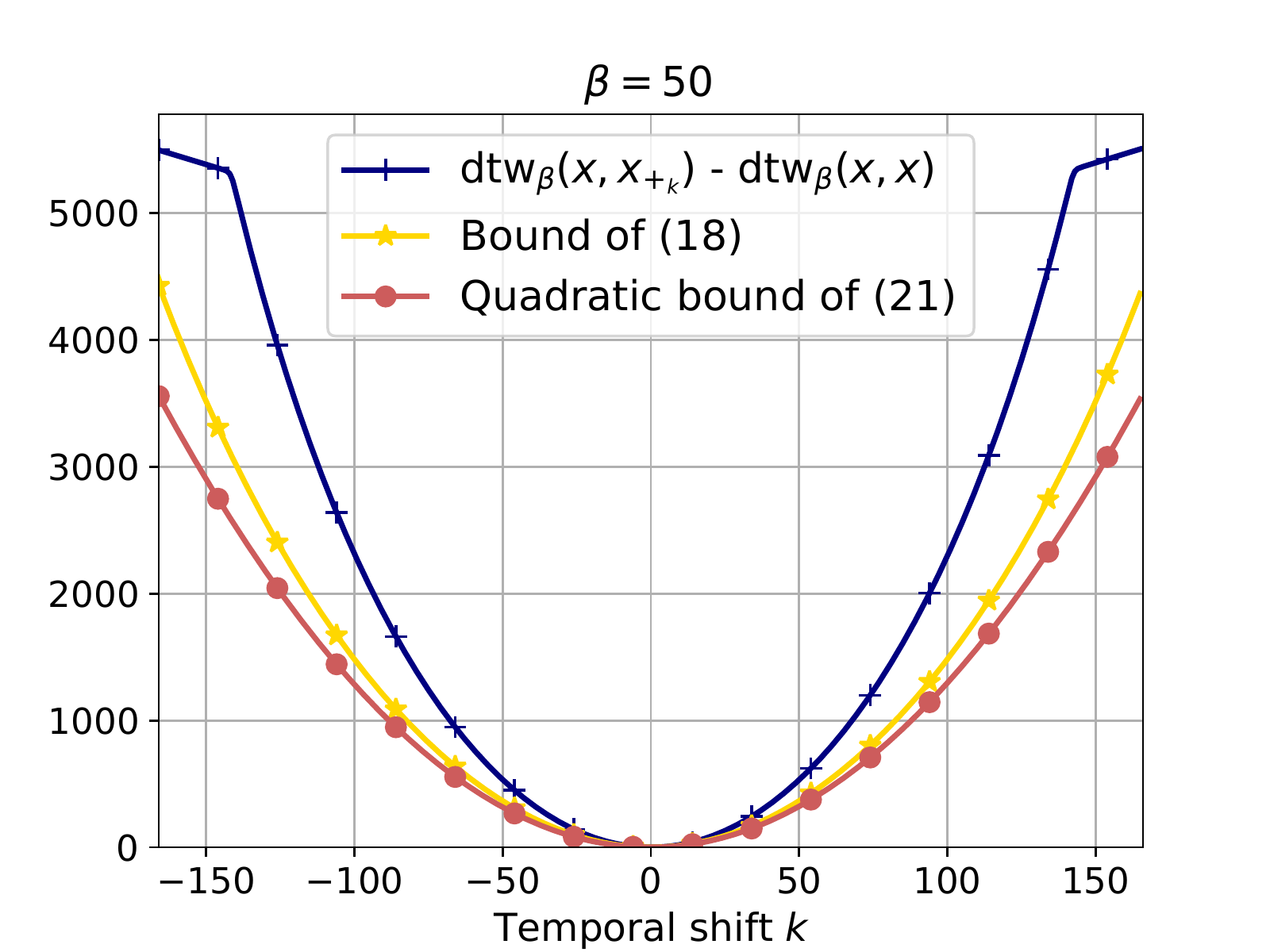}
    \end{minipage}
    \begin{minipage}{0.49\linewidth}
             \includegraphics[width=\linewidth]{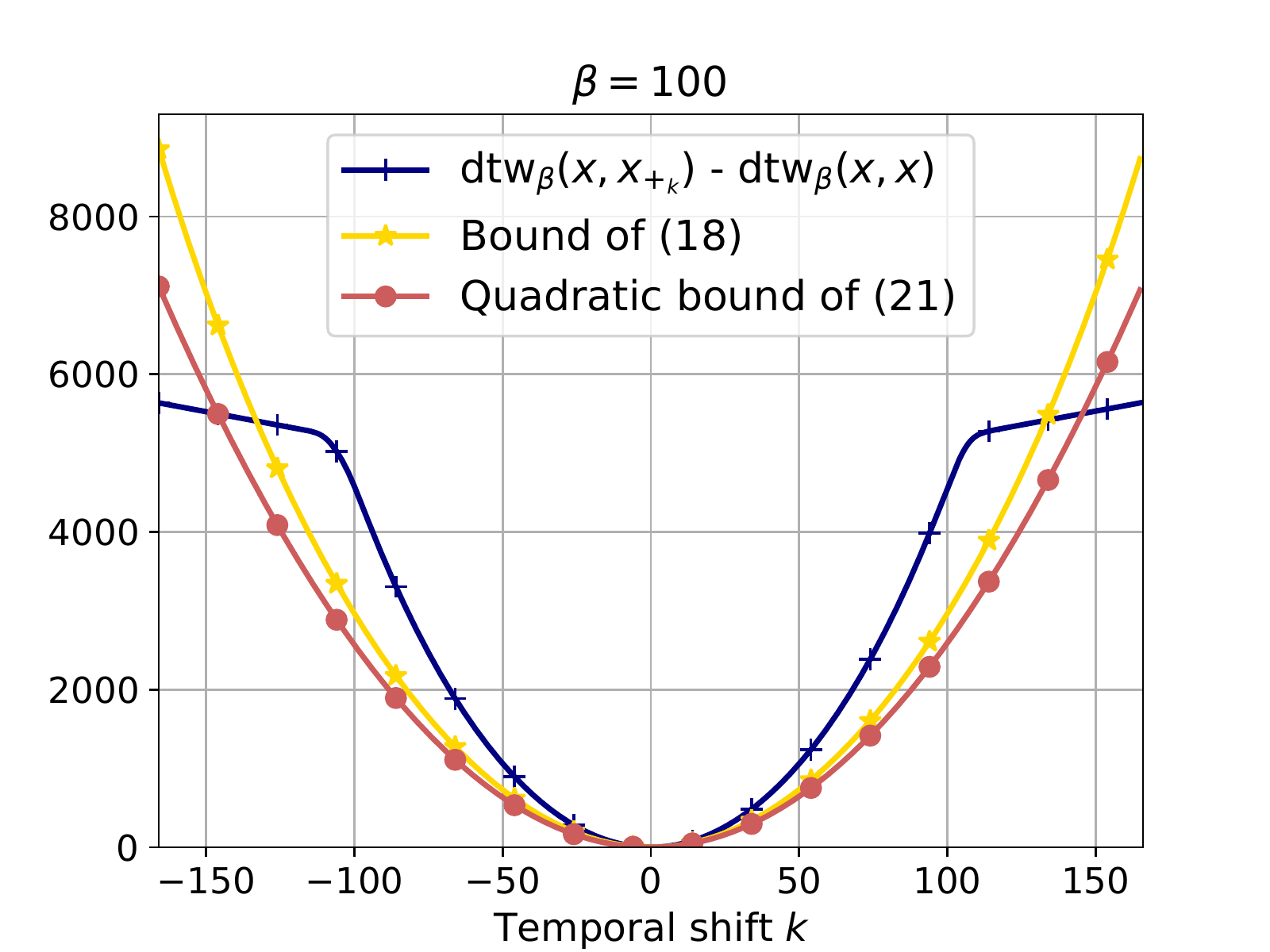}
    \end{minipage}
    \caption{Empirical evaluation of the obtained theoretical bounds for various values of $\beta$}
    \label{f:morebounds}
\end{figure}
\subsection{Brain imaging}
The time series realizations are defined on the surface of the brain which is modeled as a triangulated mesh of 642 vertices. We compute the squared ground metric M on the the mesh using Floyd-Warshall's algorithm and normalize it by its median. This normalization is standard in several optimal transport applications~\citep{otbook} and allows to scale the entropy hyperparameter $\varepsilon$ to the dimension of the data. We set $\varepsilon = 10 / 642$ and $\gamma = 1$ given by the heuristic proposed by~\citet{janati19}. Figure \ref{f:tsne-meg-all} shows additional t-SNE embeddings with different values of $\beta$.

\begin{figure}[t]
\centering
\includegraphics[width=0.5\linewidth]{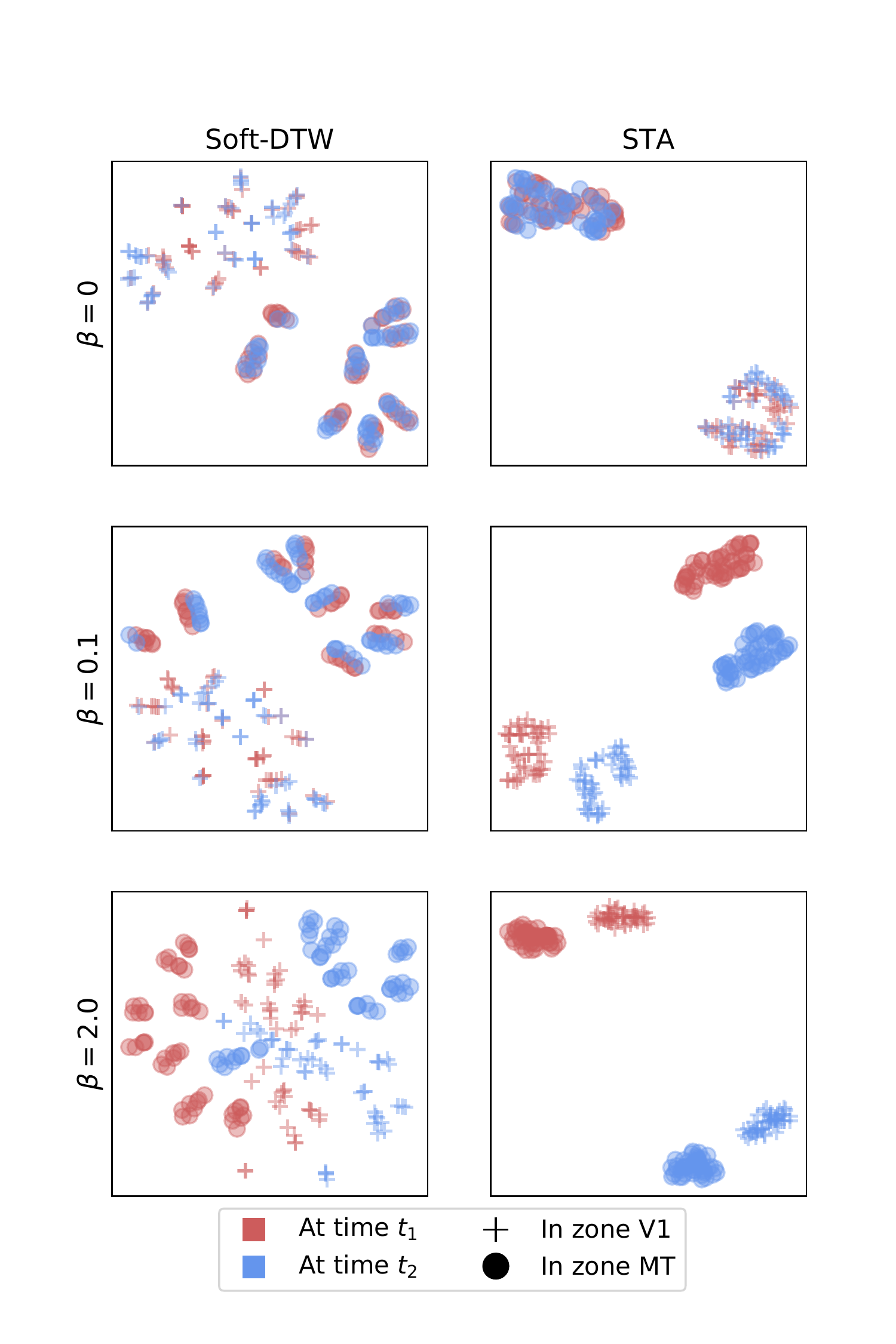}
\caption{tSNE embeddings of the data.
STA (proposed) captures spatial variability. Increasing $\beta$ helps capture more temporal variability. \label{f:tsne-meg-all}}
\end{figure}

\subsection{handwritten letters}
The raw handwritten letters data consist of (x, y) coordinates of the trajectory of the pen. The data include between 100 and 205 strokes -- time point -- for each sample. The preprocessing we performed consisted of creating the images of the cumulated trajectories and rescaling them in order to fit into a (64 $\times$ 64) 2D grid. Smoothing the data spatially. Figure \ref{f:letters} shows more examples of the processed data.
The time series realizations are defined on a 2D grid 
We compute the squared ground metric M on the the mesh using Floyd-Warshall's algorithm and normalize it by its median. This normalization is standard in several optimal transport applications~\citep{otbook} and allows to scale the entropy hyperparameter $\varepsilon$ to the dimension of the data. We set $\varepsilon = 10 / 642$ and $\gamma = 1$ given by the heuristic proposed by~\citet{janati19}. Figure \ref{f:tsne-chars-all} shows additional t-SNE embeddings with different values of $\beta$.

\begin{figure}[t]
\centering
\includegraphics[width=0.5\linewidth]{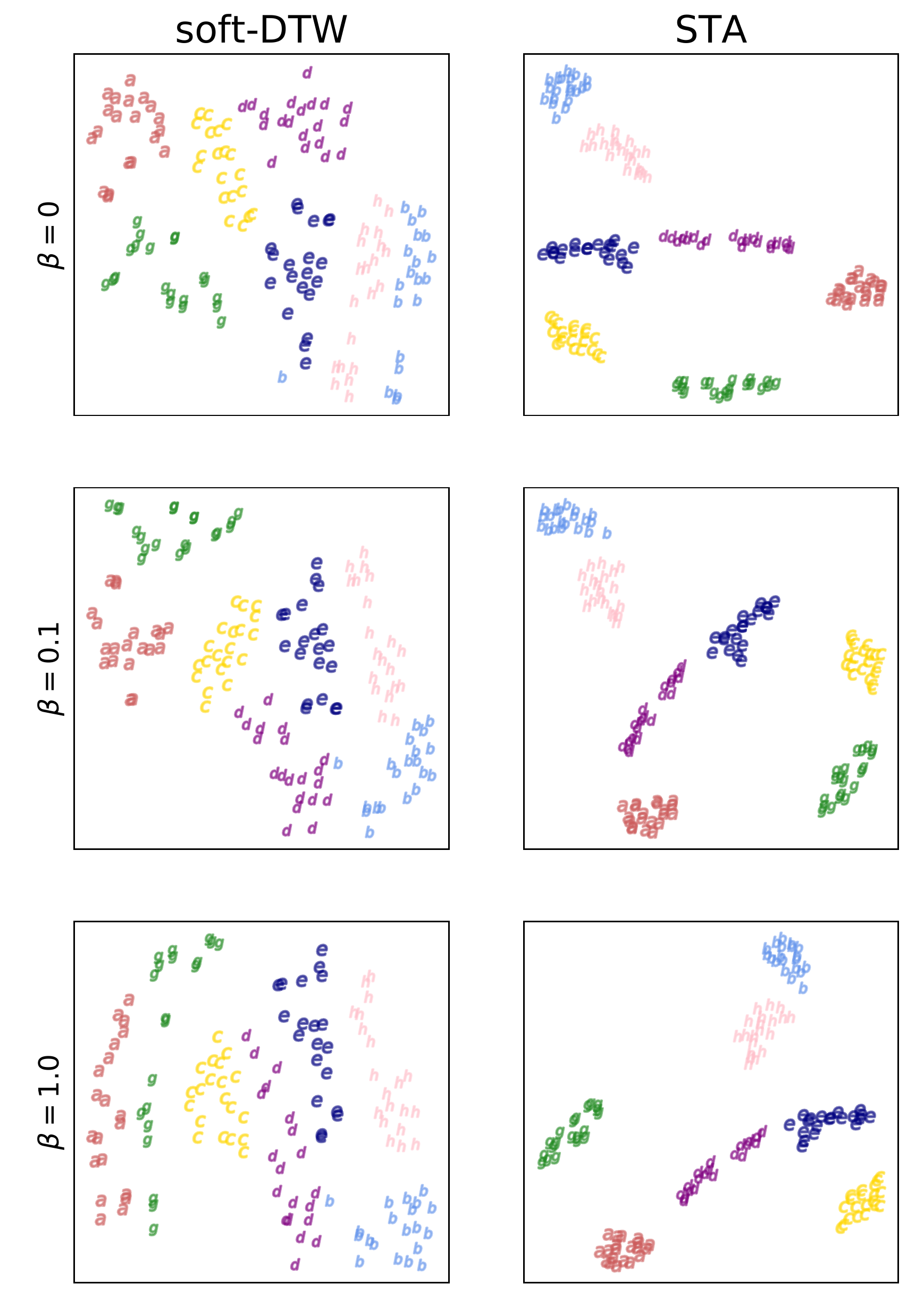}
\caption{tSNE embeddings of the data.
STA (proposed) captures spatial variability. \label{f:tsne-chars-all}}
\end{figure}

\begin{figure}[t]
\centering
\includegraphics[width=\linewidth]{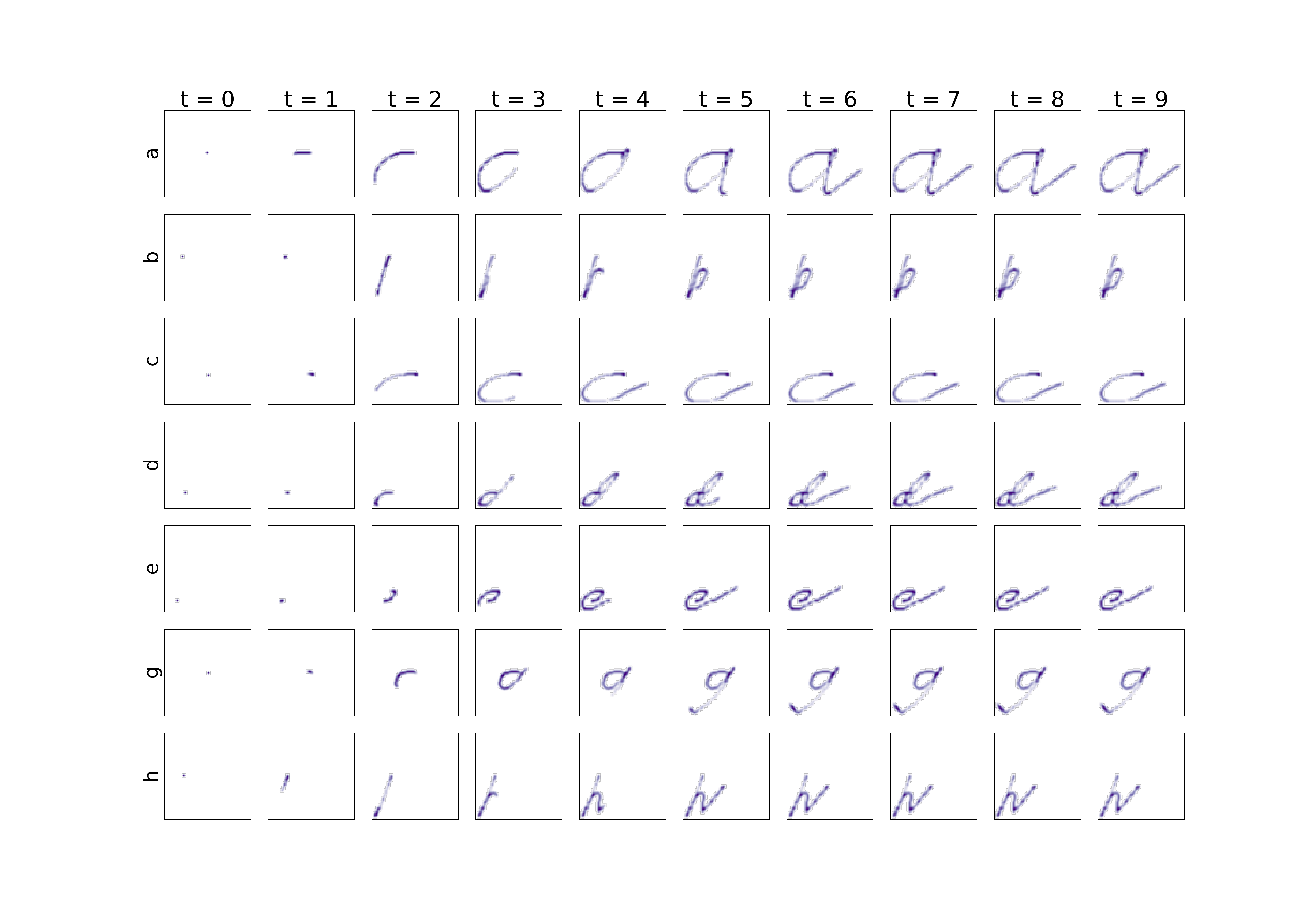}
\caption{An example of each handwritten letter in the dataset \label{f:letters}}
\end{figure}

\section{Code}

The code is provided in the supplementary materials folder. Please follow the guidelines in the README file to reproduce all experiments.

}{}

\end{document}